\documentclass[twoside,11pt]{article}
\usepackage{jmlr2e}

\usepackage{color}
\usepackage{amssymb}
\usepackage{amsmath}
\usepackage{graphicx}
\usepackage{cite}
\usepackage{bm}
\usepackage{import}
\usepackage{multirow}
\usepackage{hhline}
\usepackage{parcolumns}
\usepackage{todo}
\usepackage{longtable}
\usepackage[utf8x]{inputenc}
\usepackage[T1]{fontenc}
\usepackage[ruled,vlined,linesnumbered]{algorithm2e}
\usepackage{comment}
\usepackage[multiple]{footmisc}
\usepackage{subcaption}

\newcommand{\beq}{\begin{equation}}
\newcommand{\eeq}{\end{equation}}
\graphicspath{{./img/}}
\newcommand{\mx}[1]{\ensuremath{\boldsymbol{\mathrm{{#1}}}}}   % Simple matrix format.
\newcommand{\mxapp}[1]{\ensuremath{\hat{\mx{#1}}}}

\newcommand{\mklaren}{\texttt{mklaren}}
\newcommand{\csi}{\texttt{csi}}
\newcommand{\icd}{\texttt{icd}}
\newcommand{\ctalign}{\texttt{align}}
\newcommand{\ctalignf}{\texttt{alignf}}
\newcommand{\ctalignfc}{\texttt{alignfc}}
\newcommand{\uniform}{\texttt{uniform}}
\newcommand{\nystrom}{\texttt{Nystr\"om}}

\usepackage{url}
\urldef{\mailsa}\path|{martin.strazar, tomaz.curk}@fri.uni-lj.si|

\begin{document}

\title{Learning the kernel matrix by predictive low-rank approximations}
\jmlrheading{1}{2016}{1-48}{4/00}{10/00}{Martin Stra\v{z}ar \and Toma\v{z} Curk }

\begin{comment}
\author{\IEEEauthorblockN{Martin Stražar}
\IEEEauthorblockA{Faculty of Computer and Information Science \\ University of Ljubljana,\\ Večna pot 113, 1000 Ljubljana, Slovenia \\ Email: martin.strazar@fri.uni-lj.si}
\and
\IEEEauthorblockN{Tomaž Curk}
\IEEEauthorblockA{Faculty of Computer and Information Science \\ University of Ljubljana,\\ Večna pot 113, 1000 Ljubljana, Slovenia \\ Email: tomaz.curk@fri.uni-lj.si}
}
\end{comment}

\jmlrheading{1}{2016}{1-48}{4/00}{10/00}{}

% Short headings should be running head and authors last names

\editor{}

\author{\name Martin Stra\v{z}ar \email martin.strazar@fri.uni-lj.si \\
       \addr Bioinformatics Laboratory, Faculty of Computer and Information Science\\
       University of Ljubljana\\
       Ve\v{c}na pot 113, 1000 Ljubljana, Slovenia
        \AND
       \name Toma\v{z} Curk \email tomaz.curk@fri.uni-lj.si \\
       \addr Bioinformatics Laboratory, Faculty of Computer and Information Science\\
       University of Ljubljana\\
       Ve\v{c}na pot 113, 1000 Ljubljana, Slovenia}

% \institute{Faculty of Computer and Information Science \\ University of Ljubljana, Večna pot 113, 1000 Ljubljana, Slovenia. \mailsa \\}

\maketitle

\begin{abstract}Efficient and accurate low-rank approximations of multiple
data sources are essential in the era of big data. The scaling of kernel-based
learning algorithms to large datasets is limited by the $O(n^2)$ computation
and storage complexity of the full kernel matrix, which is required by most of
the recent kernel learning algorithms. 

We present the \mklaren\ algorithm to approximate multiple
kernel matrices learn a regression model, which is entirely based on
geometrical concepts. The algorithm does not require access to full kernel
matrices yet it accounts for the correlations between all kernels. It uses
Incomplete Cholesky decomposition, where pivot selection is based on
least-angle regression in the combined, low-dimensional feature space. The
algorithm has linear complexity in the number of data points and kernels. When
explicit feature space induced by the kernel can be constructed, a mapping from the dual
to the primal Ridge regression weights is used for model interpretation.

The \mklaren\ algorithm was tested on eight standard regression datasets. It outperforms
contemporary kernel matrix approximation approaches when learning with multiple
kernels. It identifies relevant kernels, achieving highest explained
variance than other multiple kernel learning methods for the same number of
iterations. Test accuracy, equivalent to the one using full kernel matrices,
was achieved with at significantly lower approximation ranks. A difference in
run times of two orders of magnitude was observed when either the number of
samples or kernels exceeds 3000. \end{abstract}

%%  \keywords{Multiple kernel learning, low-rank matrix factorization, Cholesky decomposition, least-angle regression.}

\section{Introduction}
% Kernel methods, linear SVM, applications, Vapnik, Cortes, Cristianini,
% Shawe-Taylor, Smola and Scholkopf
Kernel methods are popular in machine learning as they model relations between
objects in feature spaces of arbitrary, even infinite
dimension~\citep{scholkopf2002learning}. 
Kernels are \emph{inner product} functions and provide means to rich
representations, which is useful for learning in domains not associated to
vector spaces, such as structured objects, strings or trees. 
Computation of inner product values for all pairs of data points to obtain the
kernel matrix requires large computation and storage, which scales as $O(n^2)$
in the number of data instances. Kernel approximations are thus indispensable
when learning on large datasets and can be classified in two groups:
approximations of the \emph{kernel function} or approximations of the
\emph{kernel matrix}.

% Approximating the kernel function
Direct approximation of the kernel function can achieve significant performance
gains. A large body of work relies on approximating the frequently used
Gaussian kernel, which has a rapidly decaying eigenspectrum, as proved by the
Bochner's theorem and the concept of random features~\citep{Pennington,
Szabo2015, Rahimi2007}.  Recently, the property of matrices generated by the
Gaussian kernel were further exploited to achieve sublinear complexity in the
approximation rank~\citep{Yang2014,Si2014,Le2013a}. In another line of work,
low-dimensional features can be derived for translation-invariant kernels based
on their Fourier transform~\citep{Vedaldi2012}. These methods currently present
the most space- and time- efficient approximations, but are limited to kernels
of particular functional forms.

% Approximating the kernel matrix
Approximations of the kernel matrix are applicable to any symmetric
positive-definite matrix even if the underlying kernel function is unknown.
Such approximations, termed \emph{data dependent}, can be obtained using the
eigenvalue decomposition, by selecting a subset of data points (the Nystr\"om
method) or by using the Cholesky decomposition, which minimizes the divergence
between the original matrix and its low-rank
approximation~\citep{rudi2015less,Xu2015,Li2015,Gittens2013,Williams2001, Fine2001}. However, these
methods are unsupervised, they disregard side information, e.g., the target
variables. Predictive decompositions use the target variables to obtain a
supervised low-rank approximation via the Cholesky with Side
Information~\citep{Bach2005} or they minimize the Bregman divergence
measures~\citep{Kulis2009}.   The effects of kernel matrix approximation has
been discussed in context of sparse Gaussian processes \citep{Qui2005}, where
the approximation leads to degenerate Gaussian process.  Learning the inducing
points is equivalent to learning the pivots in matrix decompositions, but can
be replaced by optimizing over the whole input domain~\citep{Snelson2006,
Wilson2015}, with necessarily continuous domain.  \citet{Cao2015} relax this
limitation with a hybrid approach to kernel function and inducing set optimization.
All methods listed so far operate on single kernels. This presents a
limitation, since the choice of optimal kernel for a given learning task is
often non-trivial.

% Learning the kernel
Similarly to kernel (matrix) approximation, approaches learning the optimal
kernel for a given task (dependent on the data) can be classified to i)
learning the kernel (covariance) function or ii) learning the kernel matrix.
Learning the kernel function is possible in continuous domains, where kernel
hyperparameters are optimized to match the training
data~\citep{Mohsenzadeh2015,bishop2006pattern}.  Alternatively, kernel
functions can be learned via Fourier transforms from corresponding
power-spectrums~\citep{Gal2015, Wilson2013}.

% Approximating the kernel matrix with side information
Multiple kernel learning (MKL) methods learn the optimal weighted sum of given
kernel matrices with respect to the target variables, such as class
labels~\citep{Gonen2011}. Different kernels can thus be used to model the data
and their relative importance is assessed via the predictive accuracy, offering
insights into the problem domain.  Depending on the user-defined constraints,
the resulting optimization problems are quadratic (QP) or semidefinite programs
(SDP), assuming the complete knowledge of the kernel matrices.
\citet{Cortes2012} solve a QP on centered kernel matrices, which corresponds to
centering the data in the original input space.  Low-rank approximations have
been used for MKL, e.g., by performing Gram-Schmidt orthogonalization and
subsequently MKL \citep{Kandola2002}. In a recent study, the combined kernel
matrix is learned via efficient generalized Forward-backward algorithm, however
assuming that low-rank approximations are available
beforehand~\citep{Rakotomamonjy2014}. G\"{o}nen et al. develop a Bayesian
treatment for joint dimensionality reduction and classification, solved via
gradient descent \citep{Gonen2010} or variational
approximation~\citep{Kaski2014}, while assuming access to full kernel matrices
and not exploiting their symmetric structure.

% Hint of the results (contributions)
In this work, we propose a joint treatment of low-rank kernel approximations
and MKL. We assume an input of: i) a set of objects with corresponding
continuous target variables and a ii) set of kernels that define inner products
on the same objects.  We designed \mklaren, a greedy algorithm that couples
Incomplete Cholesky Decomposition and Least-angle regression to learn a
low-rank approximation of the combined kernel matrix. 
Our innovative approach to pivot column selection is closely associated to the
selection of feature vectors in least-angle regression (LAR)~\citep{Efron2004}.
At each step, the method keeps a current estimate of the regression model
within the span of the current approximation. In comparison to existing
methods, it has the following two advantages. First, the method is aware of
multiple kernel functions. In each step, the next pivot column to be added is
chosen greedily from all remaining pivot columns from all kernels.  Kernels
that give more information about the current residual are thus more likely to
be selected. In contrast to methods that assume access to the complete kernel
matrices, the importance of a kernel is estimated at the time of its
approximation. Also, this is different from performing the decomposition for
each kernel $k_q$ independently and subsequently determining kernels weights. 
Second, the criterion only considers the gain with
respect to the current regression residual; the notion of kernel matrix
approximation error is completely abolished. Even though accurate approximation
is proportional to the similarity of the model using the full kernel
matrix~\citep{cortes2010two}, it was recently shown that i) the expected
generalization error is related to maximal marginal degrees of freedom rather
than the approximation error and ii) empirically, low-rank approximation can
lead to a regularization-like effect~\citep{Bach2012}. Nevertheless, the
residual approximation error is guaranteed to monotonically decrease by
definition of Cholesky decompositions.

When explicit feature space representation is available for kernels, the
relation between primal and dual regression weights is used for model
interpretation. In contrast to MKL algorithms, which rely on convex
optimization or Bayesian methods, our approach relies on geometrical principles
solely, leading to a straightforward algorithm with low computational
complexity in the number of data points and kernels.

% Algorithm design Transductive setting
A common assumption when applying matrix approximation or MKL is that the
resulting decomposition can only be applied in \emph{transductive
learning}~\citep{Zhang2012, Lanckriet2004a}, i.e., the test data samples are
included in the model training phase. We apply the lemma on the uniqueness of
the low-rank approximation for a fixed active set to relate the Incomplete
Cholesky decomposition and the Nystr\"om method. With this we circumvent the
limitation to the transductive setting, infer low-rank representation of
arbitrary test data point, enabling out-of-sample prediction.

% List of results
The predictive performance, run times and model interpretation were evaluated
empirically on multiple benchmark regression datasets. The provided
implementation of \mklaren\ compared favorably against related low-rank kernel
matrix approximation and state-of-the-art MKL approaches. Additionally, we
isolate the effect of low-rank approximation and compare the method with full
kernel matrix MKL methods on a very large set of rank-one kernels.

% Section contents
The article is structured as follows. The \mklaren\ algorithm with pivot column
updates with the LAR-based selection criterion is presented in
Section~\ref{s:mklaren_overview}.  Auxilliary results regarding out-of-sample
prediction, model interpretation and computational complexity analysis are
given in Section~\ref{s:auxilliary}.  Experimental evaluation is presented in
Section~\ref{s:results}. Description of the Least-angle regression method is
given in the Appendix. The algorithm implementation and code to reproduce 
the presented experiments is available at \url{https://github.com/mstrazar/mklaren}.

\section{Multiple kernel learning with least-angle regression}
\label{s:mklaren_overview}

% Initial definition of variables
Let $\{\mx{x}_1, \mx{x}_2, ..., \mx{x}_n\}$ be a set of points in a Hilbert
space $\mathcal{X}$ of arbitrary dimension, associated with targets $\mx{y} \in
\mathbb{R}^n$.  Let the Hilbert spaces $\mathcal{X}_1, \mathcal{X}_2, ...,
\mathcal{X}_p$ be isomorphic to $\mathcal{X}$ and endowed with respective inner
product (kernel) functions $k_1, k_2, ... k_p$. The kernels $k_q$ are positive
definite and map from $\mathcal{X}_q \times \mathcal{X}_q$ to $\mathbb{R}$.
Hence, a data point $\mx{x}_i \in \mathcal{X}_q$ can be represented in multiple
inner product spaces, which can be related to different data views or
representations.  Evaluating $k_q$ for each pair of $\mx{x}_i$ determines a
kernel matrix $\mx{K}_q \in \mathbb{R}^{n \times n}$. The goal of a predictive
(supervised) approximation algorithm is to learn the corresponding low-rank
approximations $\mx{G}_1, \mx{G}_2, ..., \mx{G}_p$, where $\mx{G}_q \in
\mathbb{R}^{n \times j_q}$, $K = \sum_q j_q < n$, using additional information on the targets.  
In the context of regression, the regression line $\mx{\mu} \in \mathbb{R}^n$
is learned simultaneously with the approximations, as their construction
depends on the residual vector $\mx{r} = \mx{y} - \mx{\mu}$.

% Short summary of this section
The \texttt{mklaren} algorithm simultaneously learns low-rank approximations of
kernel matrices $\mx{K}_q$ associated to each kernel function $k_q$ and the
regression line $\mx{\mu}$.  It uses Incomplete Cholesky Decomposition (ICD) to
iteratively construct each $\mx{G}_q$. At each iteration, a kernel $k_q$ and a pivot
column $i \in {1, 2, ..., n}$ are chosen using a heuristic that evaluates the
explained information on the residual $\mx{r}$ for each potential new column of
$\mx{G}_q$.  This is achieved by using least-angle regression
in the space spanned by the previously selected (normalized and
centered) pivot columns of all $\mx{G}_q$.

% Roadmap to section
The high-level pseudo code of the \mklaren\ algorithm is shown in
Algorithm~\ref{a:mklaren}, and its steps are described in detail in
the following subsections.

\subsection{Simultaneous Incomplete Cholesky decompositions}

We start with the description of Incomplete Cholesky Decomposition (ICD) of a single
kernel matrix, and later extend it to simultaneous decomposition of multiple
kernels.  A kernel matrix \mx{K} is approximated with a Cholesky factor
$\mx{G}$. The ICD is a family of methods that produce a finite sequence of
matrices $\mx{G}^{(1)}, \mx{G}^{(2)}, ..., \mx{G}^{(j)}$, such
that 

\beq    
\mx{G}^{(j)} \mx{G}^{(j)\ T} \rightarrow \mx{K} \text{ as } j
\rightarrow n.  
\label{e:icd_sequence}
\eeq

Initially, $\mx{G}$ is initialized to $\mx{0}$. A diagonal vector representing
the lower-bound on the approximation gain is initialized as ${\mx{d} =
\text{diag}(\mx{K})}$ The active set $\mathcal{A} = \emptyset$, keeping track
of selected pivot columns. At iteration $j$, a pivot $i$ is selected from the
remaining set $\mathcal{J} = \{1, 2, ...,n \} \setminus \mathcal{A}$ and its
pivot column $\mx{G}(:, j) \leftarrow \mx{g}_i$ is computed as follows:

\beq
\begin{split}
    \mx{G}(i, j) &= \sqrt{\mx{d}(i)} \\
    \mx{G}(\mathcal{J}, j) &= \frac{1}{\mx{G}(i, j)} 
    \bigg(\mx{K}(\mathcal{J}, i) - \sum_{l=1}^{j-1}
    \mx{G}(\mathcal{J}, l)\mx{G}(i, l) \bigg) \\
\label{e:icd}
\end{split}
\eeq

Importantly, only the information on one column of $\mx{K}$ is required at each
iteration and $\mx{G}\mx{G}^T$ need never be computed explicitly. The selected pivot is
added to the active set, the counter $j$ and the diagonal vector are updated:

\beq
\begin{split}
    \mx{d}      &\leftarrow \mx{d} - \mx{g}_j^2 \\
    \mathcal{A} &\leftarrow \mathcal{A} \cup \{i\}\\
    j           &\leftarrow j + 1
\label{e:icd_2}
\end{split}
\eeq

\begin{figure}[t]
\centering
\includegraphics[width=0.99\textwidth]{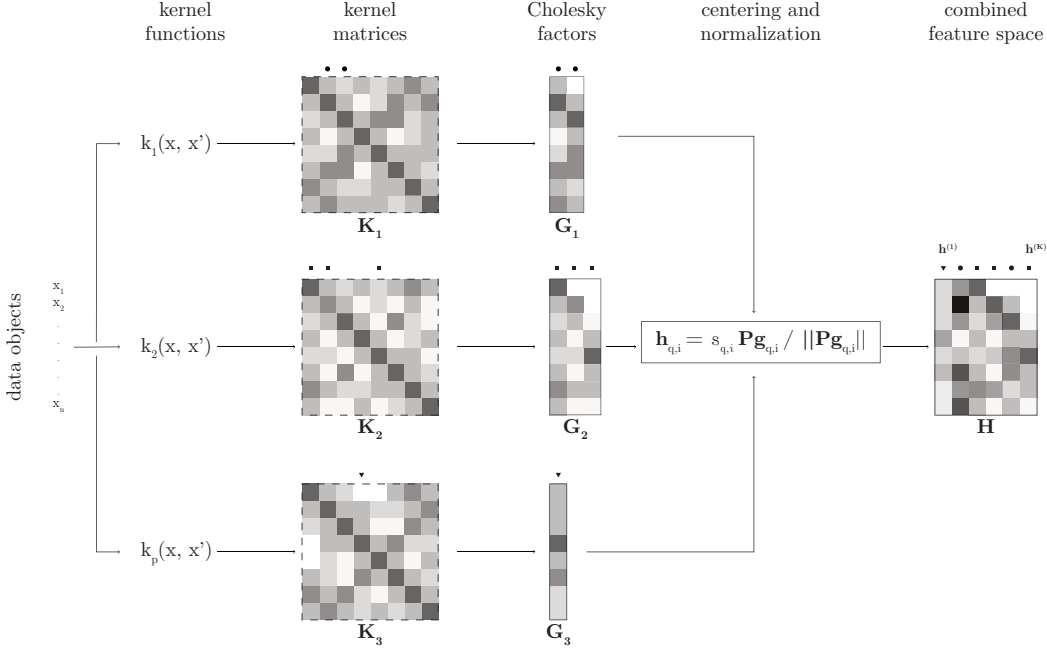}
\caption{Overview of variables included in the hypothetical model using three
kernels, $q\in\{1, 2, 3\}$.  Kernel matrices in dashed values are never
computed explicitly. The markers \emph{circle}, \emph{rectangle} and
\emph{triangle} represent the selected pivot columns for kernels $1, 2$ and $3$
respectively.}
\label{f:overview-1}
\end{figure}

In the case of multiple, $p$ kernels, each kernel function $k_q$, $q \in {1,
2,..., p}$ determines a corresponding $\mx{K}_q$, which is approximated with
Cholesky factors $\mx{G}_q$. An example scenario is depicted on Fig.~\ref{f:overview-1}.

The set of all Cholesky factors $\mx{G}_q$ is used to construct a
\emph{combined feature matrix} to be used for least-angle regression, as
follows. At any point, assume the existence of the residual vector $\mx{r}$, to
be constructed in Section~\ref{ss:lar_selection}.  For each selected pivot
column $\mx{g}_{q, i}$ in kernel $q$ define the following transformation.

\beq 
    \mx{h}_{q, i} \leftarrow s_{q, i} \mx{P} \mx{g}_{q, i} / \| \mx{P} \mx{g}_{q, i} \|
    \label{e:h}
\eeq

where the operator \mx{P} is the centering projection $\mx{P} = (\mx{I} -
\frac{\mx{1}\mx{1}^T}{n})$ and $s_{q, i}$ is the sign of the correlation
$(\mx{P}\mx{g}_i)^T\mx{r}$. Each $\mx{h}_{q, i}$ is normalized and makes an
angle at most 90 degrees with the residual \mx{r}.  The set of columns
$\mx{h}_{q, i}$ span the \emph{combined feature space}, equivalent to any matrix
$\mx{H} \in \mathbb{R}^{n \times \sum_q j_q}$ containing this same set of
columns (in any order).

\beq
\text{span}(\mx{H}) = \text{span} \big(\{ \mx{h}_{1,1}, \mx{h}_{1,2}...,\mx{h}_{1,j_1}, \mx{h}_{2,1},..., \mx{h}_{2,j_2},\mx{h}_{p, 1},...,\mx{h}_{p, j_p} \} \big)
\label{def:bh}
\eeq

The Fig.~\ref{f:overview-1} shows one such matrix. Note that applying the operator
$\mx{P}$ is equivalent to centering the positive semidefinite matrix
$\mx{H}\mx{H}^T$, which represents the \emph{combined kernel}.

Least-angle regression is used to iteratively select the next pivot column and
thus determine the order of columns in \mx{H} while simultaneously updating the
regression line. The next kernel $q$ and pivot $j$ are selected from all
remaining sets $\mathcal{J}_q$, based on the current residual~$\mx{r}$. The
corresponding pivot column $\mx{g}_{q, i}$ is computed using the Cholesky step
in Eq.~\ref{e:icd} and added to $\mx{G}_q$. At any iteration, each $\mx{G}_q$ may
contain a different number of columns $j_q$ as their selection depends on the
relevance for explaining the residual.

\subsection{Pivot selection based on Least-angle regression}
\label{ss:lar_selection}

Least-angle regression (LAR) is an \emph{active set method}, for feature subset
selection in linear regression  (see Appendix and \citet{Efron2004} for
thorough description). Here, we propose an idea based on the LAR column
selection to determine the next pivot column to be added to any of the
$\mx{G}_q$ and consequently to combined feature matrix $\mx{H}$.

The original LAR method assumes availability of all variables representing the
covariates (column vectors) in the sample data matrix. In our case, however,
this matrix is $\mx{H}$ and is constructed iteratively. The adaptation of
the LAR-based column selection is non-trivial, since the exact values of the
new columns $\mx{g}_{q, i}$ and $\mx{h}_{q, i}$ are unknown at selection time. 

This section describes a method to construct $\mx{H}$ given the columns
$\mx{h}_{q, i}$ and learn $\mx{\mu} \in \text{span}(\mx{H})$.  In favor of
clarity we assume (only in this section) that values of all $\mx{h}_{q, i}$ are
known and describe the ordering of $\mx{h}_{q, i}$ in $\mx{H}$. The problem
of unknown candidate pivot column values is postponed to Section~\ref{s:lookahead}.

The matrix $\mx{H}$ is initialized to $\mx{0}$. 
The regression line $\mx{\mu}$ and the residual $\mx{r}$ are initialized

\beq
\mx{\mu} = \mx{0} \text{ and } \mx{r} = \mx{y}, \text{ assuming w.l.g. } \mx{1}^T\mx{y} = 0.
\eeq

By construction, $\|\mx{h}_{q, i}\| = 1$ and $\mx{1}^T\mx{h}_{q, i} = 0$ for all $q, i$.
The $h_{q, i}$ will be added to $\mx{H}$ in a defined ordering
\beq
    \mx{H}(:, l) \leftarrow \mx{h}_{q, i} = \mx{h}^{(l)} \text{ for } l = 1, 2,...\sum_q j_q
    \label{e:h_definition}  
\eeq

where a unique kernel, pivot pair $q, i$ is selected for each position $l$.
The ordering depends on the correlation with the residual  $c_l = \mx{r}^T
\mx{h}^{(l)}$. Therefore, the Cholesky factors $\mx{G}_q$ containing pivot
columns with more information on the current residual are selected preferably.

\begin{figure}
\centering
\begin{subfigure}[b]{0.6\textwidth}
    \centering
    \includegraphics[width=1.05\textwidth]{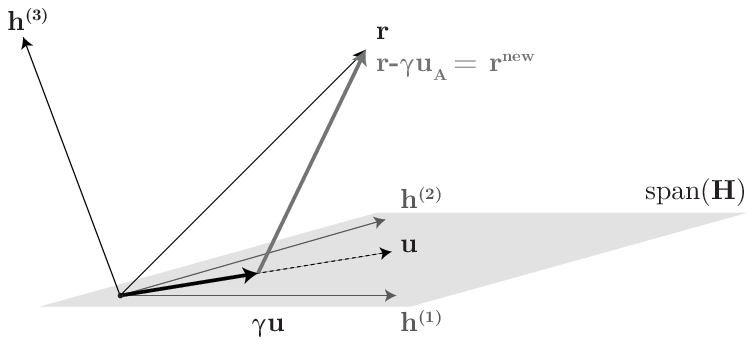}
    \label{f:overview-2a}
    \caption{}
\end{subfigure}
\begin{subfigure}[b]{0.3\textwidth}
    \centering
    \includegraphics[width=0.5\textwidth]{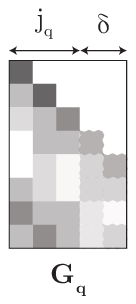}
    \label{f:overview-2b}
    \caption{}
\end{subfigure}

\caption{
a) Updating the regression line within the combined feature matrix
\mx{H} containing two vectors $\mx{h}^{(1)}$ and $\mx{h}^{(2)}$. 
The residual is $\mx{r} = \mx{y} - \mx{\mu}$,
where $\mx{\mu} \in \text{span}(\mx{h}^{(1)})$ and 
$\measuredangle(\mx{r}, \mx{h}_2) = \measuredangle(\mx{r}, \mx{h}_1)$. 
The new residual $\mx{r}^{new}$
upon selection of $\mx{h}^{(2)}$ is obtained by 
adding $\gamma\mx{u}$ to $\mx{\mu}$ and updating $\mx{r}$ accordingly. 
The step size $\gamma$
is increased until some new vector $\mx{h}^{(3)}$ will have the same correlation (angle) with $\mx{r}^{new}$
as both $\mx{h}^{(1)}$ and $\mx{h}^{(2)}$, i.e., 
$\measuredangle(\mx{r}^{new}, \mx{h}_3) = \measuredangle(\mx{r}^{new}, \mx{h}_2) = \measuredangle(\mx{r}^{new}, \mx{h}_1)$ .
b) Schematic representation of selected $j_q$ pivot columns and $\delta$ look-ahead columns. 
}
\label{f:overview-2}

\end{figure}

The column selection procedure is depicted in Fig.~\ref{f:overview-2}a and is
defined as follows.  At iteration $l=1$, the first vector $\mx{h}^{(1)}$ is chosen to maximize correlation 
$\mx{h}^{(1)} = \text{max}_{m = 1...\sum_q j_q} c_m = \mx{r}^T \mx{h}^{(m)}$. This $\mx{h}^{(1)}$
is added to $\mx{H}(:, 1) = \mx{h}^{(1)}$. 

At each iteration $l$, \mx{H} contains $l$ columns $\mx{h}^{(1)}, \mx{h}^{(2)},
..., \mx{h}^{(l)}$. By elementary linear algebra, there exist the \emph{bisector}
$\mx{u}$, having $\|\mx{u}\| = 1$ and making equal angles, less than 90
degrees, between the residual $\mx{r}$ and vectors vectors currently in
$\mx{H}$.  Updating the regression line $\mx{\mu}$ along direction $\mx{u}$ and
the residual $r$

\beq
    \mx{\mu}^{\text{new}} = \mx{\mu} + \gamma \mx{u}  \hspace{5mm}
    \mx{r}^{\text{new}}   = \mx{r} - \gamma \mx{u} 
\label{e:lar_update}
\eeq

causes the correlations $c_l = \mx{r}^T \mx{h}^{(l)}$ to change equally
for all $\mx{h}^{(l)}$ in \mx{H}, for an arbitrary step size $\gamma \in \mathbb{R}$. 
The value $\gamma$ is set such that some new column $h^{(l+1)}$ not in \mx{H}
will have the same correlation to $r^{(new)}$ as all the columns
already in $\mx{H}$:
\beq
  \measuredangle(\mx{r}^{new}, \mx{h}^{(1)}) = ... 
= \measuredangle(\mx{r}^{new}, \mx{h}^{(l)}) = 
  \measuredangle(\mx{r}^{new}, \mx{h}^{(l+1)})
\eeq

% Determination of step size
The step size $\gamma$ and $ \mx{h}^{(l+1)}$ are selected as follows. 
Define the following quantities 

\beq
C = \text{max}_{\{l | h^{(l)} \in \mx{H}\}} c_l \hspace{5mm} A = (\mx{1}^T\mx{H}\mx{1})^{-1/2}.
\eeq
Then, 
\beq
\begin{split}
\gamma &= \text{min}^+_{\{m | \mx{h}^{(m)} \notin \mx{H}\}} \bigg\{\frac{C -c_m}{A - a_m}, \frac{C + c_m}{A + a_m} \bigg\} , \text{where} \\
c_m &= \mx{r}^T\mx{h}^{(m)} \\
a_m &= \mx{u}^T\mx{h}^{(m)}.
\end{split}
\label{e:lar_gamma}
\eeq

Here, $\text{min}^+$ is the minimum over positive arguments for each choice of
$m$.  The selected column vector $\mx{h}^{(m)}$ is the minimizer of
Eq.~\ref{e:lar_gamma} and is inserted at the $l+1$-th position in \mx{H},
$\mx{H}(:, l+1) = \mx{h}^{(m)}$. For the last column vector (as there are no
further column vectors to chose from) the step size simplifies to $\gamma = C /
A$, yielding the ordinary least-squares solution for $\mx{H}$ and \mx{y}.

% Linker text to next section
The mentioned problem in our case is that the exact values of all potential
pivot columns $\mx{g}_{q, i}$ not in $\mx{G}_q$ and its corresponding $\mx{h}_{q, i}$
are unknown.  Explicit calculation of all columns using the Cholesky step in
Eq.~\ref{e:icd} would yield quadratic computational complexity, as their values
are dependent on all previously selected pivots. The issue is addressed by
using approximations $\hat{\mx{g}}_{q, i}$ and $\mxapp{h}_{q, i}$ that are
less expensive to compute, as described in the following section.

\subsection{Look-ahead decompositions}
\label{s:lookahead}

The selection of a new column vector $\mx{h}^{(m)}$ to be added to the combined
feature matrix \mx{H} and its corresponding $\mx{h}_{g, i}$, $\mx{g}_{q, i}$ is
based only on the values $a_m$, $c_m$ in Eq.~\ref{e:lar_gamma}. Instead of
explicitly calculating each candidate $\mx{g}_{q, i}$ for all $q, i$ at each
iteration, we use an approximate column vector $\hat{\mx{g}}_{q, i}$. The
approach uses a similar idea to look-ahead (information) columns in
\citep{Cao2015,Bach2005}.

Consider the kernel matrix $\mx{K}_q$, its current
Cholesky factor $\mx{G}_q$ and active set $\mathcal{A}_q$. 
By definition of ICD in Eq.~\ref{e:icd}, the values of a
candidate pivot column $\mx{g}_{q, i}$ at step $j$ and pivot $i \notin \mathcal{A}_q$ are:
\beq
\mx{g}_{q, i} = \frac{(\mx{K}_q - \sum_{l=1}^{j_q}\mx{G}_q(:, l)\mx{G}_q(:, l)^T)(:, i)}{\sqrt{\mx{d}_q(i)}}
\label{e:chol_step}
\eeq

The main computation cost in the above definition is the computation 
of a rank-$n$ kernel matrix $\mx{K}_q$ for each $m$. Instead,
$\delta$ look-ahead columns are used to get a \emph{look-ahead approximation} 
$\mx{L}_q = \mx{G}_q(:, j_q+\delta)\mx{G}_q(:, j_q+\delta)^T$ (Fig.~\ref{f:overview-2}b).
This defines approximate values  $\mxapp{g}_{q, i}$:

\beq
\begin{split}
\mxapp{g}_{q, i} &= \frac{(\mx{L}_q- \sum_{l=1}^{j-1}\mx{G}_q(:, l)\mx{G}_q(:, l)^T)(:, i)}{\sqrt{\mx{d}_q(i)}} \\
            &= \frac{\mx{G}_q(:,\ j_q{+}1{:}j_q{+}\delta) \  \mx{G}_q^T (j_q{+}1{:}j_q{+}\delta,\ i)}{\sqrt{\mx{d}_q(i)}}
\label{e:chol_step_lookahead}
\end{split}
\eeq

\newpage
Given $\mxapp{g}_{q, i}$ and consequently $\mxapp{h}_{q, i}$, consider the computation of $\hat{c}_{q, i}$:

\beq
\hat{c}_{q, i} = \mx{r}^T\mxapp{h}_{q, i} =  \frac{| (\mx{P} \mxapp{g}_{q, i})^T \mx{r}| }{\|\mx{P} \mxapp{g}_{q, i}\|} \\
\label{e:app_c}
\eeq

Inserting $\mxapp{g}_{q, i}$ as in Eq.~\ref{e:chol_step_lookahead}, the
denominator $1/\sqrt{\mx{d}_q(i)}$  cancels out. The norm $\|\mx{P}
\mxapp{g}_{q, i}\|$ can be computed as:
\beq
\begin{split}
       \|\mx{P} \mxapp{g}_{q, i}\|^2 
                    &= \bigg( \mx{P} \  \mx{G}_q(:,\ j_q{+}1{:}j_q{+}\delta)\  \mx{G}_q^T (j_q{+}1{:}j_q{+}\delta,\ i) \bigg)^T \bigg( \mx{P} \ \mx{G}_q(:,\ j_q{+}1{:}j_q{+}\delta) \ \mx{G}_q^T (j_q{+}1{:}j_q{+}\delta,\ i) \bigg) \\
                    &= \mx{G}_q(i, :) \bigg( \mx{G}_q^T\mx{G}_q(j_q{+}1{:}j_q{+}\delta,\ j_q{+}1{:}j_q{+}\delta) - \mx{G}_q^T\mx{1}\mx{1}^T\mx{G}_q(j_q{+}1{:}j_q{+}\delta,\ j_q{+}1{:}j_q{+}\delta)
                         \bigg) \mx{G}_q(i, :)^T
\end{split}
\eeq

Similarly, dot product with the residual is computed as:
\beq
\begin{split}
    (\mx{P} \mxapp{g}_{q, i})^T\mx{r} &= \mx{r}^T \bigg( \mx{P} \ \mx{G}_q(:,\ j_q{+}1{:}j_q{+}\delta) \mx{G}_q^T(j_q{+}1{:}j_q{+}\delta,\ i) \bigg)^T = \\
         &=  \bigg( \mx{r}^T\mx{G}_q(:,\ j_q{+}1{:}j_q{+}\delta) - \mx{r}^T \mx{1}\mx{1}^T \mx{G}_q(:,\ j_q{+}1{:}j_q{+}\delta) \bigg) \mx{G}_q(j_q{+}1{:}j_q{+}\delta,\ i)
\end{split}
\label{e:app_c_dot}
\eeq

Computation of $\hat{a}_{q, i}$ is analogous.  Correctly ordering the order of
computation yields the computational complexity $O(\delta^2)$ per column.  Note
that matrices $\mx{G}_q^T\mx{G}_q$, $\mx{G}_q^T\mx{1}\mx{1}^T\mx{G}_q$,
$\mx{r}^T\mx{G}_q$, $\mx{r}^T \mx{1}\mx{1}^T \mx{G}_q$ are the same for all columns
(independent of $i$) and need to be computed only once per iteration.

% Recomputing j:j+delta
The values $\hat{a}_{q, i}$ and $\hat{c}_{q, i}$ can be computed efficiently for all kernel matrices
and enable the selection of next kernel, pivot column pair $q, i$ to be added
to $\mx{G}_q$ and consequently $\mx{H}$. After selecting $q, i$ a Cholesky step
in performed (Eq.~\ref{e:icd}) to compute the exact $\mx{g}_{q, i}$ and 
\beq
    \mx{G}_q(:, j) \leftarrow \mx{g}_{q, i}
    \label{e:icd_new_update}
\eeq

The computation of a new column renders the look-ahead columns in $\mx{G}_q$ at indices $j_q{+}1{:}j_q{+}\delta$
invalid. After applying Eq.~\ref{e:icd_new_update}, all columns at indices $j_q{+}1{:}j_q{+}\delta$
are recomputed using standard Cholesky step with pivot selection based on current maximal value
in $\mx{d}_q$ at a cost $O(n\delta^2)$.

% Recomputing gamma
The exact values of $\mx{g}_{q, i}$ and $\mx{h}_{q, i}$ determine
$\mx{h}^{(m)} $ to be added to \mx{H} and enables the correct computation of  $a_m$, $c_m$ 
and step size $\gamma$ in Eq.~\ref{e:lar_gamma}. The regression line $\mx{\mu}$ and the
residual $\mx{r}$ can be correctly updated according to Eq.~\ref{e:lar_update}.

\subsection{The \mklaren\ algorithm}
\label{s:wrapup}

The steps described in previous sections complete the \mklaren\ algorithm.
Given a sample of $n$ data objects with a targets $\mx{y}$ and $p$
kernel functions, the user specifies three additional parameters: the maximum
rank $K$ of combined feature matrix, number of look-ahead columns $\delta$ and
$L_2$ regularization parameter $\lambda$ (constraining \mx{\mu}, discussed in
Section~\ref{s:regularization}). 

The variables related to regression line ($\mx{\mu}$, residual $\mx{r}$ and
bisector $\mx{u}$) and individual decompositions $\mx{G}_q$ (active sets
$\mathcal{A}$, column counters $j_q$) are initialized in
lines~\ref{l:init_start}-\ref{l:init_end}. Each $\mx{G}_p$ is initialized using standard ICD with $\delta$ look-ahead
columns, as described in Section~\ref{s:lookahead} (line~\ref{l:lookahead}).

The main loop is executed for $K$ iterations, until the sum of selected pivot
colums equals $\sum_q j_q = K$, where at each iteration a kernel $k_q$ and a
pivot column $i$, $i \notin \mathcal{A}_q$ are selected and added to $\mx{G}_q$
and consequently the combined feature matrix \mx{H}. For each kernel $k_q$ and
each pivot $i \notin \mathcal{A}_q$, $\hat{a}_{q, i}$ and $\hat{c}_{q, i}$ are
computed. Based on these approximated values, the kernel $k_q$
and pivot $i$ are selected.
Given the optimal $k_q$ and pivot $i$, the pivot column $\mx{g}_{q, i}$ and
$\mx{h}_{q, i}$ are computed. The new pivot column $\mx{g}_{q, i}$ is added to $\mx{G}_q(:,
j_q)$, $j_q$ is incremented and the $\delta$ columns at $\mx{G}_q(:,
j_q{+}1{:}j_q{+}\delta)$ are recomputed using standard ICD
(lines~\ref{l:selection_start}-\ref{l:selection_end}).

Having computed the exact $\mx{g}_{q, i}$, the true values $a_{q, i}$ and $c_{q, i}$
can be computed and the regression line $\mx{\mu}$ and the residual
are updated (lines~\ref{l:lar_start}-\ref{l:lar_end}). 

The regression coefficients \mx{\beta} solving $\mx{H}\mx{\beta} = \mx{\mu}$,
required for out-of-sample prediction, can be obtained by
constructing \mx{H} and solving a linear system discussed in Section~\ref{s:weights}
(line~\ref{l:beta}).

\begin{algorithm}
\SetKwData{Variables}{variables}

\KwIn{ \\
\hspace{5mm} $ \{\mx{x}_1, \mx{x}_2, ..., \mx{x}_n\}$ set of objects in $\mathcal{X}$, \\ 
 \hspace{5mm} $k_1, k_2, ... k_p$ kernel functions on $\mathcal{X} \times \mathcal{X}$, \\
 \hspace{5mm} $\mx{y} \in \mathbb{R}^n$ regression targets, with $\mx{1}^T\mx{y} = 0$, \\
 \hspace{5mm} $K$ maximum total rank, \\
 \hspace{5mm} $\delta$ number of look-ahead columns,  \\
 \hspace{5mm} $\lambda$ regularization parameter. \\
} 

\BlankLine
\KwResult{ \\
 \hspace{5mm} $\mx{G}_1 \in \mathbb{R}^{n \times j_1}, 
 \mx{G}_2 \in \mathbb{R}^{n \times j_2}, ...
 \mx{G}_p \in \mathbb{R}^{n \times j_p} $, \\ \hspace{10mm} Cholesky factors, \\
 \hspace{5mm} $\mx{H} \in \mathbb{R}^{n \times K}$ combined feature space, \\ 
 \hspace{5mm} $\mathcal{A}_1, \mathcal{A}_2, ..., \mathcal{A}_p$ active sets of pivot indices, \\
 \hspace{5mm} $\mx{\mu} \in \mathbb{R}^{n}$ regression line on the training set, \\
 \hspace{5mm} $\mx{\beta} \in \mathbb{R}^K$ regression coefficients. \\
}

\BlankLine
\BlankLine

Initialize: \label{l:init_start} \\
\hspace{5mm} $\mx{H} = \mx{0}$, 

\hspace{5mm} residual $\mx{r} = \mx{y}$, 

\hspace{5mm} bisector $\mx{u} = \mx{0}$, 

\hspace{5mm} regression line $\mx{\mu} = \mx{0}$,

\hspace{5mm} active sets $\mathcal{A}_q = \emptyset$ and counters $j_q = 0$ for $q \in \{1, ..., p\}$ \label{l:init_end}.

% ider ook-ahead steps
\BlankLine
    Compute standard Cholesky Decompositions with $\delta$ look-ahead columns for $\mx{G}_1, \mx{G}_2, ..., \mx{G}_p$.
    \label{l:lookahead}

% Main loop
\BlankLine
\BlankLine
\While{$\sum_{q} j_q < K$}{

    \BlankLine
    
    Compute $\hat{a}_{q, i}$ and $\hat{c}_{q, i}$ for each kernel $q$ and pivot $i \notin A_{q}$ (Eq.~\ref{e:app_c}) 
    \label{l:selection_start} \\
    Select $q, i$ based on the minimum in Eq.~\ref{e:lar_gamma} \\ 
    Compute $\mx{g}_{q, i}$ (Eq.~\ref{e:icd}) and $\mx{h}_{q, i}$ (Eq.~\ref{e:h})\\
    \hspace{5mm} $\mx{G}_q(:, j_q) \leftarrow \mx{g}_{q, i}$  \\
    \hspace{5mm} $\mx{H}(:, \sum_q j_q) \leftarrow \mx{h}_{q, i}$ \\
    \hspace{5mm} $j_q \leftarrow j_q + 1, \mathcal{A}_q \leftarrow \mathcal{A}_q \cup \{i\}$ \\
    \BlankLine
    Recompute $\mx{G}_q(:,j_q{+}1{:}j_q{+}\delta)$ using standard ICD \label{l:selection_end} \\

    \BlankLine
    \BlankLine
    Compute true $a_{q, i}$ and $c_{q, i}$ (Eq.~\ref{e:lar_gamma}) \label{l:lar_start}\\
    Compute the bisector \mx{u} of columns in \mx{H} except $h_{q, i}$ (Eq.~\ref{e:bisector}) \\
    Compute $\gamma$ for $\mx{h}^{(m)} = \mx{h}_{q, i}$ (Eq.~\ref{e:lar_gamma}) and update \\
    \hspace{5mm}    $\mx{\mu} \leftarrow \mx{\mu} + \gamma\mx{u}$ \\
    \hspace{5mm}    $\mx{r}   \leftarrow \mx{r} - \gamma\mx{u}$ \label{l:lar_end}\\
}

\BlankLine
Solve linear system $\mx{H}\mx{\beta} = \mx{\mu}$ for $\mx{\beta}$ using Eq.~\ref{e:beta_qr}.\label{l:beta}

\caption{The \mklaren\ algorithm pseudocode.}
\label{a:mklaren}
\end{algorithm}

\section{Auxiliary theoretical results}
\label{s:auxilliary}

This section presents auxiliary theoretical results required for out-of-sample
prediction (Sections~\ref{s:weights}-\ref{s:outofsample}), model
interpretation using the relation between primal and dual regression coefficients
(Section~\ref{s:dual_coef}), $L_2$ regularizaion (Section~\ref{s:regularization}), 
and computational complexity (Section~\ref{s:complexity}).

\subsection{Computing the regression coefficients}
\label{s:weights}
The regression coefficients $\mx{\beta} \in \mathbb{R}^K$ are computed from the
regression line $\mx{\mu}$ and the combined feature space \mx{H} as defined in
Eq.~\ref{def:bh}.
using the relation
\beq
\mx{H}\mx{\beta} = \mx{\mu}  \implies \mx{\beta} = (\mx{R}^T\mx{R})^{-1}\mx{Q}^T \mx{\mu},
\label{e:beta_qr}
\eeq

where $\mx{H} = \mx{QR}$ is the thin QR decomposition~\citet{golub2012matrix}.

\subsection{Out-of-sample prediction}
\label{s:outofsample}

Inference of Cholesky factors corresponding to test (unseen) samples is possible without
explicitly repeating the Cholesky steps.  The coefficients $\mx{\beta}$ are
then used to predict the responses for new samples.  To simplify notation, we
show the approach for one kernel matrix and its corresponding Cholesky factors,
while the computation for multiple kernels is analogous.

\textbf{Nystr\"om approximation}. Let $\mathcal{A} = \{i_1, i_2, ..., i_j\}$ be
an arbitrary active set of pivot indices. The Nystr\"om aproximation~\citep{Williams2001} of the
kernel matrix \mx{K} is defined as follows:

\beq
    \mx{L} = \mx{K}(:,\mathcal{A})\mx{K}(\mathcal{A},\mathcal{A})^{-1}\mx{K}(:,\mathcal{A})^T
\eeq

The construction of $\mathcal{A}$ crucially influences
the prediction performance. Note that \mklaren\ defines a method to construct
$\mathcal{A}$.

% Let $\mx{G}\mx{G}^T$ be an (Incomplete) Cholesky decomposition using the same
% pivot indices in the set $\mathcal{A}$. Consider the following proposition:

\vspace{3mm}
\textbf{Proposition.} \emph{The Incomplete Cholesky decomposition with pivots
$\mathcal{A} = \{i_1, i_2, ..., i_j\}$ yields the same approximation as the
Nystr\"om approximation using the active set $\mathcal{A}$.}
\beq
  \mx{L} = \mx{G}\mx{G}^T 
         = \mx{K}(:,\mathcal{A})\mx{K}(\mathcal{A},\mathcal{A})^{-1}\mx{K}(:,\mathcal{A})^T
\eeq

\vspace{3mm}

The proof follows directly from~\citet{Bach2005}, Proposition 1. There exists an unique matrix
\mx{L} that is: \emph{(i) symmetric,  (ii) has the column space spanned by $\mx{K}(:,
\mathcal{A})$ and (iii) $\mx{L}(:, \mathcal{A}) = \mx{K}(:, \mathcal{A})$.} It
follows that both Incomplete Cholesky decomposition and the Nystr\"om approximation
result in the same approximation matrix $\mx{L}$.

\vspace{3mm}
\textbf{Corollary.} Let $\mx{G} \in \mathbb{R}^{n \times K}$ be the Cholesky factors 
obtained on the training set $\{\mx{x}_1, \mx{x}_2, ... \mx{x}_n\}$ using pivots indices $\mathcal{A}$. 
Let  $\mx{K}(*, A)$ be the values of the kernel function $k(\mx{x}^*,
\mx{x}_i)$ evaluated for all test samples $\mx{x}^*$ and training samples in the active set $\mx{x}_i$, for $i \in \mathcal{A}$.
The Cholesky factors $\mx{G}^* \in \mathbb{R}^{t \times K}$ for test samples $\{ \mx{x}_1^*,
\mx{x}_2^*, ...\mx{x}^*_{t} \}$ are inferred using the linear transform 
$\mx{T} = \mx{K}(\mathcal{A}, \mathcal{A}) \mx{K}(\mathcal{A}, :)\mx{G}(\mx{G}^T\mx{G})^{-1} $. 
\vspace{3mm}

\beq
\begin{split}
    \mx{G}^{*}\mx{G}^{T} &= \mx{K}(*, \mathcal{A})  \mx{K}(\mathcal{A},\mathcal{A})^{-1} \mx{K}(\mathcal{A}, :) \\  
    & \implies \\
    \mx{G}^{*} &= \mx{K}(*, A) \mx{K}(\mathcal{A}, \mathcal{A}) \mx{K}(\mathcal{A}, :)\mx{G}(\mx{G}^T\mx{G})^{-1} \\
               &= \mx{K}(*, \mathcal{A}) \mx{T}
\end{split}
\label{e:nystrom_cholesky}
\eeq

\hfill $\blacksquare$
\vspace{2mm}

The matrix $\mx{T} \in \mathbb{R}^{K{\times}K}$ is inexpensive to compute and
can be stored permanently after the training phase. Hence, the Cholesky factors
$\mx{G}^*$ are computed from the inner product between the test and the active
sets $\mx{K}(\mathcal{A}, *)$. The combined feature matrix $\mx{H}^{*} \in \mathbb{R}^{t \times K}$ and the predictions $\mx{\mu}^* \in \mathbb{R}^t$ are obtained after
centering and normalization against the training Cholesky factors \mx{G}:
\beq
\begin{split}
    \mx{H}^{*}(:, j) &= \frac{\mx{G}^{*}(:, j) - \mx{P}\mx{G}(:, j)}{\| \mx{P}\mx{G}(:, j) \|} \text{ for } j \in 1...K\\           
    \mx{\mu}^* &= \mx{H}^{*} \mx{\beta}
\end{split}
\eeq
where $\mx{P} = \mx{I} - \frac{\mx{11}^T}{n}$ and $\mx{\beta}$ is defined in Eq.~\ref{e:beta_qr}.

\subsection{Computing the dual coefficients}
\label{s:dual_coef}

Regardless of using the approximation to kernels, a limited form of model interpretation is still 
possible for a certain class of kernels.  Again, we show the approach for one kernel matrix and the combined feature 
matrix \mx{H} while the computation for multiple kernels is analogous.

Kernel ridge regression is often stated in terms of dual coefficients $\mx{\alpha} \in \mathbb{R}^n$,
satisfying the relation:

\beq
\mx{H}^T \mx{\alpha} =  \mx{\beta}
\eeq

This is an overdetermined system of equations. The vector \mx{\alpha} with
minimal norm can be obtained by solving the following least-norm problem:
\beq
\begin{split}
    \text{minimize }     &\|\mx{\alpha}\|_2 \\
    \text{subject to }   &\mx{H}^T \mx{\alpha} = \mx{\beta}
\end{split}
\eeq

The problem has an analytical solution equal to
\beq
   \mx{\alpha} = \mx{H} (\mx{H}^T\mx{H})^{-1}\mx{\beta}
\eeq

Obtaining dual coefficients $\mx{\alpha}$ can be useful if the range of the explicit feature map
induced by a kernel $k$ is finite, such that $k(x, x') = \Phi(\mx{x})\Phi(\mx{x}')$, $\Phi: \mathcal{X} \mapsto \mathbb{R}^P$ which is the case for
linear, polynomial, and various string kernels~\citep{sonnenburg2005large}. An interpretation of regression
coefficients in the range of~$\Phi$, $\mx{\beta}_{\Phi} \in \mathbb{R}^{P}$ is obtained by computing the matrix $\mx{\Phi} \in \mathbb{R}^{n \times P}$
for the training set and considering
\beq
    \mx{\beta}_{\Phi} =\mx{\Phi}^T \mx{\alpha}.
\eeq

Moreover, if the vector \mx{\alpha} is sparse, only the relevant portions of
$\mx{\Phi}$ need to be computed.  This condition can be enforced
by using techniques such as matching pursuit when solving for \mx{\alpha}~\citep{bach2010sparse}.

\subsection{$\ell_2$ norm regularization}
\label{s:regularization}

Regularization is achieved by constraining the norm of weights $\|\mx{\beta}\|$.
\citet{Zou2005} prove the following lemma, which shows that
$\ell_2$ regularized regression problem can be stated as ordinary least
squares using an appropriate augmentation of the data $\mx{X}, \mx{y}$.
The following lemma assumes for all $l$, $\|\mx{X}(:, l)\| = 1$, $\mx{1}^T\mx{X}(:, l) = 0$ and 
$\mx{1}^T\mx{y}= 0$. 

\vspace{2mm} \textbf{Lemma}. Define the augmented data set $\mx{X}^{\lambda}, \mx{y}^{\lambda}$
to equal

\begin{equation*} \mx{X}^{\lambda} = \sqrt{(1+\lambda)} \begin{pmatrix}\mx{X} \\
\sqrt{\lambda}\mx{I}\end{pmatrix} \end{equation*} \begin{equation*} \mx{y}^{\lambda} =
\begin{pmatrix} \mx{y} \\ \mx{0} \end{pmatrix}.  \end{equation*}

The least-squares solution of $\mx{X}^{\lambda}\mx{\beta}=\mx{y}^{\lambda}$ is then equivalent to Ridge
regression of the original data $\mx{X}, \mx{y}$ with parameter $\lambda$. For
proof, see \citet{Zou2005}. The augmented dataset can be included in LAR to
achieve the $\ell_2$-regularized solution. This is achieved by modifying the columns of the combined
feature matrix in Eq.~\ref{def:bh}:
\beq
    \mx{h}^{\lambda}_{q, i}  = \mx{P} \begin{pmatrix} \mx{P}\mx{g}_{q, i} \\ 0 \\ 0 \\ ... \\ \lambda \\ ... \\ 0 \end{pmatrix} / \| \mx{P} \begin{pmatrix} \mx{P}\mx{g}_{q, i} \\ 0 \\ 0 \\ ... \\ \lambda \\ ... \\ 0 \end{pmatrix}\|
\label{e:h_aug}
\eeq

This definition is now equivalent to performing LAR in augmented space $\mx{H}^{\lambda}$, resulting in an $\ell_2$
regularized solution for $\mx{\mu}$ after $K$ steps of the approximations. It is straightforward to modify
Eq.~\ref{e:lar_gamma}, and Eq.~\ref{e:app_c}-\ref{e:app_c_dot} for $\mx{h}^{\lambda}_{q,i}$.

\subsection{Computational complexity} 
\label{s:complexity}

The \mklaren\ algorithm scales as a linear function of both the number of data
points $n$ and kernels $p$. The computational complexity is
\beq
O(n\delta^2 + K(K^2 + np\delta^2 + n\delta^2) + nK^2 + K^3)  =
O(K^3 + npK\delta^2).
\label{e:complexity}
\eeq

The look-ahead Cholesky decompositions are standard Cholesky decompositions
with $\delta$ pivots and complexity $O(n\delta^2)$. The main loop is executed
$K$ times. The selection of kernel and pivot pairs is based on the LAR
criterion, which includes inverting $\mx{H}_A^T\mx{H}_A$ of size $K \times K$,
thus having a complexity of $K^3$. However, as each step is a rank-one
modification to $\mx{H}_A^T\mx{H}_A$, the Morrison-Sherman-Woodbury lemma on
matrix inversion~\citep{meyer2000matrix} can be used to achieve complexity
$O(K^2)$ per update. The computation of correlations with the bisector in
Eq.~\ref{e:app_c} and residuals are computed for $p$ kernels in
$O(np\delta^2)$.  Recomputation of~$\delta$ Cholesky factors requires standard
Cholesky steps of complexity $O(n\delta^2)$. The computation of the gradient step
is of the same complexity as the gradient step. Updating the regression line is
$O(n)$. The QR decomposition in Eq.~\ref{e:beta_qr} takes $O(nK^2)$ and the
computation of linear transform $\mx{T}$ in Eq.~\ref{e:nystrom_cholesky} is of
$O(K^3 + nK^2)$ complexity.

\section{Experiments}
\label{s:results}

In this section, we provide an empirical evaluation of the proposed method
on known regression datasets. We compare \mklaren\ with several well-known
low-rank matrix approximation methods: Incomplete Cholesky Decomposition (\icd,
\citep{Fine2001}), Cholesky with side Information (\csi, \citep{Bach2005}) and
the Nystr\"om method (\nystrom, \citep{Williams2001}).  

We also compare \mklaren\ with a family of state-of-the-art multiple kernel
learning methods tha use the full-kernel matrix. The comparison was performed
on a sentiment analysis data set with a large number of rank-one
kernels~\citep{Cortes2012}.

\subsection{Comparison with low-rank approximations}

\begin{table*}[b!]
\centering

\small
\begin{tabular}{|p{1.5cm}||p{2.3cm}|p{2.3cm}|p{2.1cm}|p{2.1cm}||p{2.1cm}|} 
\hline 
Dataset & \mklaren & \csi & \icd & \nystrom & \uniform\\ 
\hline 
boston & \textbf{4.393 $\pm$ 0.432} & $4.762 \pm 0.598$ & $6.703 \pm 0.354$ & $6.611 \pm 1.272$ & $3.109 \pm 0.274$\\ 
kin & \textbf{0.018 $\pm$ 0.000} & $0.025 \pm 0.003$ & $0.067 \pm 0.006$ & $0.065 \pm 0.006$ & $0.013 \pm 0.000$\\ 
pumadyn & \textbf{1.252 $\pm$ 0.032} & $1.650 \pm 0.169$ & $4.024 \pm 0.655$ & $3.882 \pm 0.803$ & $1.210 \pm 0.070$\\ 
abalone & \textbf{2.638 $\pm$ 0.116} & $2.768 \pm 0.187$ & $2.906 \pm 0.222$ & $2.939 \pm 0.197$ & $2.499 \pm 0.118$\\ 
comp & \textbf{5.288 $\pm$ 0.461} & $7.520 \pm 1.852$ & $14.111 \pm 1.123$ & $13.763 \pm 0.580$ & $0.750 \pm 0.203$\\ 
ionosphere & \textbf{0.283 $\pm$ 0.017} & $0.310 \pm 0.022$ & $0.380 \pm 0.010$ & $0.377 \pm 0.015$ & $0.292 \pm 0.025$\\ 
bank & \textbf{0.036 $\pm$ 0.001} & $0.046 \pm 0.005$ & $0.101 \pm 0.011$ & $0.128 \pm 0.010$ & $0.034 \pm 0.001$\\ 
diabetes & \textbf{54.680 $\pm$ 3.61} & $54.953 \pm 3.018$ & $63.715 \pm 5.970$ & $68.117 \pm 3.947$ & $62.142 \pm 3.991$\\ 
\hline
\end{tabular}
\vspace{2mm}

\begin{tabular}{|p{1.5cm}||p{2.3cm}|p{2.3cm}|p{2.1cm}|p{2.1cm}||p{2.1cm}|} 
\hline 
Dataset & \mklaren & \csi & \icd & \nystrom & \uniform\\ 
\hline 
boston & \textbf{3.792 $\pm$ 0.454} & $4.481 \pm 0.689$ & $5.499 \pm 0.680$ & $5.677 \pm 0.609$ & $3.109 \pm 0.274$\\ 
kin & \textbf{0.016 $\pm$ 0.001} & $0.018 \pm 0.000$ & $0.059 \pm 0.008$ & $0.054 \pm 0.009$ & $0.013 \pm 0.000$\\ 
pumadyn & \textbf{1.257 $\pm$ 0.032} & $1.268 \pm 0.035$ & $3.552 \pm 0.767$ & $3.581 \pm 0.660$ & $1.210 \pm 0.070$\\ 
abalone & \textbf{2.526 $\pm$ 0.097} & $2.591 \pm 0.111$ & $2.777 \pm 0.194$ & $2.820 \pm 0.220$ & $2.499 \pm 0.118$\\ 
comp & \textbf{3.100 $\pm$ 0.942} & $5.318 \pm 1.298$ & $12.646 \pm 0.548$ & $11.288 \pm 2.365$ & $0.750 \pm 0.203$\\ 
ionosphere & \textbf{0.234 $\pm$ 0.028} & $0.254 \pm 0.030$ & $0.341 \pm 0.012$ & $0.331 \pm 0.016$ & $0.292 \pm 0.025$\\ 
bank & \textbf{0.035 $\pm$ 0.001} & $0.036 \pm 0.001$ & $0.067 \pm 0.005$ & $0.110 \pm 0.012$ & $0.034 \pm 0.001$\\ 
diabetes & $55.580 \pm 3.634$ & \textbf{55.220 $\pm$ 3.56} & $58.793 \pm 5.606$ & $60.747 \pm 2.377$ & $62.142 \pm 3.991$\\ 
\hline
\end{tabular}

\vspace{2mm}

\begin{tabular}{|p{1.5cm}||p{2.3cm}|p{2.3cm}|p{2.1cm}|p{2.1cm}||p{2.1cm}|} 
\hline 
Dataset & \mklaren & \csi & \icd & \nystrom & \uniform\\ 
\hline 
boston & \textbf{3.493 $\pm$ 0.489} & $4.191 \pm 0.878$ & $4.657 \pm 0.664$ & $5.220 \pm 0.751$ & $3.109 \pm 0.274$\\ 
kin & \textbf{0.014 $\pm$ 0.000} & $0.018 \pm 0.000$ & $0.043 \pm 0.018$ & $0.040 \pm 0.014$ & $0.013 \pm 0.000$\\ 
pumadyn & $1.255 \pm 0.038$ & \textbf{1.251 $\pm$ 0.027} & $3.015 \pm 0.702$ & $2.448 \pm 0.742$ & $1.210 \pm 0.070$\\ 
abalone & \textbf{2.500 $\pm$ 0.110} & $2.545 \pm 0.095$ & $2.597 \pm 0.128$ & $2.702 \pm 0.159$ & $2.499 \pm 0.118$\\ 
comp & \textbf{1.330 $\pm$ 0.409} & $4.791 \pm 2.805$ & $9.845 \pm 2.085$ & $9.744 \pm 2.005$ & $0.750 \pm 0.203$\\ 
ionosphere & \textbf{0.221 $\pm$ 0.012} & $0.228 \pm 0.018$ & $0.304 \pm 0.024$ & $0.266 \pm 0.033$ & $0.292 \pm 0.025$\\ 
bank & \textbf{0.034 $\pm$ 0.002} & $0.035 \pm 0.001$ & $0.042 \pm 0.009$ & $0.101 \pm 0.020$ & $0.034 \pm 0.001$\\ 
diabetes & $55.628 \pm 3.597$ & \textbf{55.214 $\pm$ 4.03} & $56.608 \pm 4.488$ & $57.560 \pm 2.425$ & $62.142 \pm 3.991$\\ 
\hline
\end{tabular}

\vspace{2mm}

\caption{
Comparison of regression performance (RMSE) on test sets via 5-fold
cross-validation for different values of rank ($K$).  Shown in bold is the
low-rank approximation method with lowest RMSE. \textbf{Top} K=14.
\textbf{Middle} K=28. \textbf{Bottom} K=42.
}

\label{t:rmse}
\end{table*}
The main advantage of \mklaren\ over established kernel matrix
approximation methods is simultaneous approximation of multiple kernels, which
considers the current approximation to the regression line and greedily selects
the next kernel and pivot to include in the decomposition. To elucidate
this, we performed a comparison on eight known regression datasets: abalone,
bank, boston, comp-active, diabetes, ionosphere, kinematics, pumadyn\footnote{\url{http://archive.ics.uci.edu/ml/}}\footnote{\url{http://www.cs.toronto.edu/~delve/data/datasets.html}}. 

Similar to~\citet{Cortes2012}, seven Gaussian kernels with different
length scale parameters are used. The Gaussian kernel function is defined as
$k(x, y) = \text{exp}\{-\gamma\|\mx{x}-\mx{y}\|^2\}$, where the length scale
parameter $\gamma$ is in range $2^{-3}, 2^{-2}, ..., 2^0, ..., 2^3$. For
approximation methods \icd, \csi\ and \nystrom, each kernel matrix was
approximated independently using a fixed maximum rank $K$. The combined feature
space of seven kernel matrices was used with ridge regression.

For \mklaren, the approximation is defined simultaneously for all kernels and
the maximum rank was set to $7K$, i.e., seven times the maximum rank of
individual kernels used in \icd, \csi\ and \nystrom. Thus, the low-rank feature
space of all four methods had exactly the same dimension. The uniform kernel
combination (\uniform) using the full-kernel matrix was included as an
empirical lower bound.

The performance was assessed using 5-fold cross-validation as follows.
Initially up to 1000 data points were selected randomly from the dataset. For
each random split of the data set, a \emph{training set} containing 60\% of the
data was used for kernel matrix approximation and fitting the regression model.
A \emph{validation set} containing 20\% of the data was used to select the
regularization parameter $\lambda$ from range $10^{-3}, 10^{-2}, ...  10^0,
... 10^3$. The final reported performance using root mean square error (RMSE)
was obtained on the \emph{test set} with remaining 20\% of the data.  All
variables were standardized and the targets \mx{y} were centered to the mean.
The look-ahead parameter $\delta$ was set to 10 for \mklaren\ and \csi.

The results for different settings of $K$ are shown in Table~\ref{t:rmse}.
Not surprisingly, the performance of supervised \mklaren\ and \csi\ is
consistently superior to unsupervised \icd\ and \nystrom. Moreover, \mklaren\
outperforms \csi\ on the majority of regression tasks, especially at lower
values of $K$. At higher values of $K$, the difference vanishes as
all approximation methods recover sufficient information of the feature
space induced by the kernels. 

\begin{table}[ht!]
\centering
\begin{tabular}{|ll||c|c|c|c|} 
\hline 
Dataset & $n$ & \mklaren & \csi & \icd & \nystrom\\ 
\hline 
boston & 506 & \textbf{42} & 63 & $>140$ & 119\\ 
kin & 1000 & \textbf{63} & $>140$ & $>140$ & $>140$\\ 
pumadyn & 1000 & \textbf{49} & $>140$ & 56 & 98\\ 
abalone & 1000 & \textbf{21} & 28 & 35 & 49\\ 
comp & 1000 & \textbf{49} & 63 & $>140$ & $>140$\\ 
ionosphere & 351 & \textbf{14} & \textbf{14} & 42 & 35\\ 
bank & 1000 & \textbf{21} & 42 & 42 & 112\\ 
diabetes & 442 & \textbf{14} & \textbf{14} & \textbf{14} & 21\\ 
\hline
\end{tabular}
\vspace{2mm}
\caption{Comparison of minimal rank for which the RMSE differs by at most one
standard deviation to RMSE obtained with the full kernel matrices using uniform
alignment. The number of data samples is denoted by $n$. Shown in bold is the method with lowest maximal rank
 $K$ to achieve equivalent performance to \uniform.} 
\label{t:ranks}
\end{table}

It is interesting to compare the utilization of the vectors in the
low-dimensional feature space. Table~\ref{t:ranks} shows the minimal setting of
$K$ where the performance is at most one standard deviation away from the
performance obtained by \uniform. On seven out of eight datasets, \mklaren\
reaches equivalent performance to \uniform\ at the smallest setting of $K$. The
differences in ranks among all four evaluated methods in Table~\ref{t:ranks}
are statistically significant (p=0.0012, Friedman rank-sum test). Additionaly,
\mklaren\ and \csi\ difference in ranks
is statistically significant (Wilcoxon signed-rank test, p=0.03552).
On only the diabetes dataset, the
unsupervised \icd\ and \nystrom\ outperform the supervised methods at
low ranks. However, at higher setting of $K$ the performance of \csi\ and
\mklaren\ overtakes \icd\ and \nystrom\ as can be seen in Table~\ref{t:rmse}.

Overall the results confirm the utility of the greedy approach to select not
only pivots, but also the kernels to be approximated and suggest \mklaren\ to
be the method of choice when competitive performance at very low-rank feature
spaces is desired. The kernels that are not added to the decomposition can be
discarded. This point is discussed further in the next subsection.

\subsection{Comparison with MKL methods on rank-one kernels}

The comparison of \mklaren\ to multiple kernel learning methods using the full
kernel matrix is challenging as it is unrealistic to expect improved
performance with low-rank approximation methods. Although the restriction to
low-rank feature spaces may result in implicit regularization and improved
performance as a consequence, the difference in implicit
dimension of the feature space makes the comparison difficult~\citep{Bach2012}.

We focus on the ability of \mklaren\ to select from a set of kernels in a way
that takes into account the implicit correlations between the kernels. To this
end, we built on the empirical analysis of~\citet{Cortes2012}. The
mentioned reference used four well-known sentiment analysis datasets compiled
by~\citet{blitzer2007biographies}. In each dataset, the examples
are user reviews of products and the target is the product rating in a discrete
range $1..5$. The features are counts of 4000 most frequent unigrams and
bigrams in each dataset.  Each feature was represented by a rank-one kernel,
thus enabling the use of multiple kernel learning for feature selection and
explicit control over feature space dimension. The datasets contain moderate
number of examples: books ($n=5501$), electronics ($n=5901$), kitchen
($n=5149$) and dvd ($n=5118$).  The splits into \emph{training} and \emph{test}
part were readily included as a part of the data set. 

We compared \mklaren\ with three state-of-the-art multiple kernel learning
methods for comparison. All methods are based on maximizing centered kernel
alignment~\citet{Cortes2012}. The \ctalign\ method infers the kernel weights
independently, while \ctalignf\ and \ctalignfc\ consider the between-kernel
correlations when maximizing the alignment. The combined kernel learned by all
three methods was used with kernel ridge regression model. The \ctalign\ method
is linear in the number of kernels ($p$), while \ctalignf\ and \ctalignfc\ are
cubic as the include solving an unconstrained (\ctalignf) or a constrained
(\ctalignfc) QP.

\begin{figure*}
\centering
\includegraphics[width=0.95\textwidth]{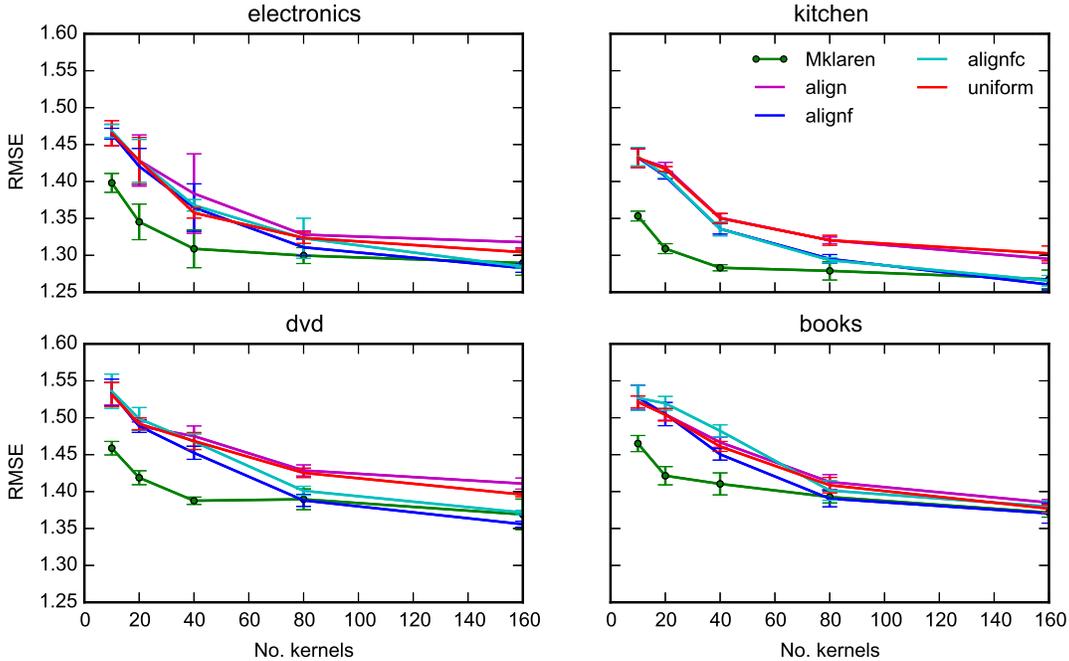}

\caption{RMSE on the test set for MKL methods. The rank $K$ is equal to the number of kernels included.}
\label{f:rmse_align_test}
\end{figure*}

When testing for different ranks $K$, the features were first filtered
according to the descending centered alignment metric for \ctalign, \ctalignf,
\ctalignfc\ prior to optimization.  When using \mklaren\ the $K$ pivot columns
were selected from the complete set of 4000 features.  The parameter $\delta$
was set to 1. Note that in this scenario, \mklaren\ is equivalent to the
original LAR algorithm, thus excluding the effect of low-rank approximation and
comparing only the kernel selection part. This way, the same dimension of the
feature space was ensured.

The performance was measured via 5-fold cross-validation. At each step, 80\% of
the \emph{training} was used for kernel matrix approximation (\mklaren) or
determining kernel weights (\ctalign, \ctalignf, \ctalignfc). The remaining
20\% of the training set was used for selecting regularization parameter
$\lambda$ from range $10^{-3}, 10^{-2}, ... 10^3$ and the final performance was
reported on the \emph{test} set using RMSE.

The results for different settings of $K$ are shown in
Fig.~\ref{f:rmse_align_test}. For low settings of $K$, \mklaren\ outperforms
all four other MKL methods that assume full kernel-matrices. The performance of
\mklaren\ at K=$40$ is within one standard deviation of best performance using
160 features using any of the methods, showing that the greedy kernel and pivot
selection criterion considers implicit correlations between kernels. 

\begin{figure}
\centering
\includegraphics[width=0.5\textwidth]{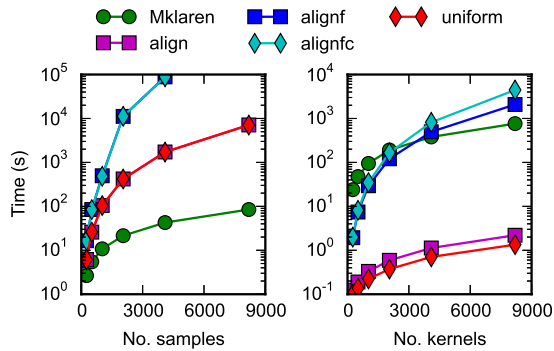}
\caption{Comparison of running times. (left) Time versus number of samples on a
synthetic dataset with $P=10$ kernels. (right) Time versus number of kernels on
\texttt{books} training dataset, $n=4000$ samples.} \label{f:time}
\end{figure}

However, there is an important difference in computational complexity.  Note
that \mklaren\ is linear in the number of kernels $p$, which presents a
practical advantage over \ctalignf\ and \ctalignfc. The comparison of methods'
implementation run times is shown on Fig.~\ref{f:time}. 

We compared the methods run times on a synthetic dataset with $p=10$
Gaussian kernels differing in parameters, rank $K=40$ and variable $n$. Since
the methods (\ctalign, \ctalignf, \ctalignfc, \uniform) require the computation
of the whole kernel matrix, \mklaren\ was significanlty more efficient (up to 3
orders of magnitude with $n$=4000 samples).

The experiments with varying number of kernels were performed on the
\texttt{books} dataset. The centered kernel alignment value can be computed
efficiently due to the usage of rank-one linear kernels, without explicit
computation of the outer product. This proves very efficient for \uniform\ and
\ctalign\ methods where the weights are computed independently. While the
\mklaren\ method is also linear in the number kernels ($p$), it accounts for
the in-between kernel correlations. The overhead in calculating low-rank
approximations is beneficiary when the number of kernels exceeds 2000. Thus,
effect for accounting of between-kernel correlations is achieved at a
significantly computational lower cost.

\begin{figure*}[t]
\centering
\includegraphics[width=0.95\textwidth]{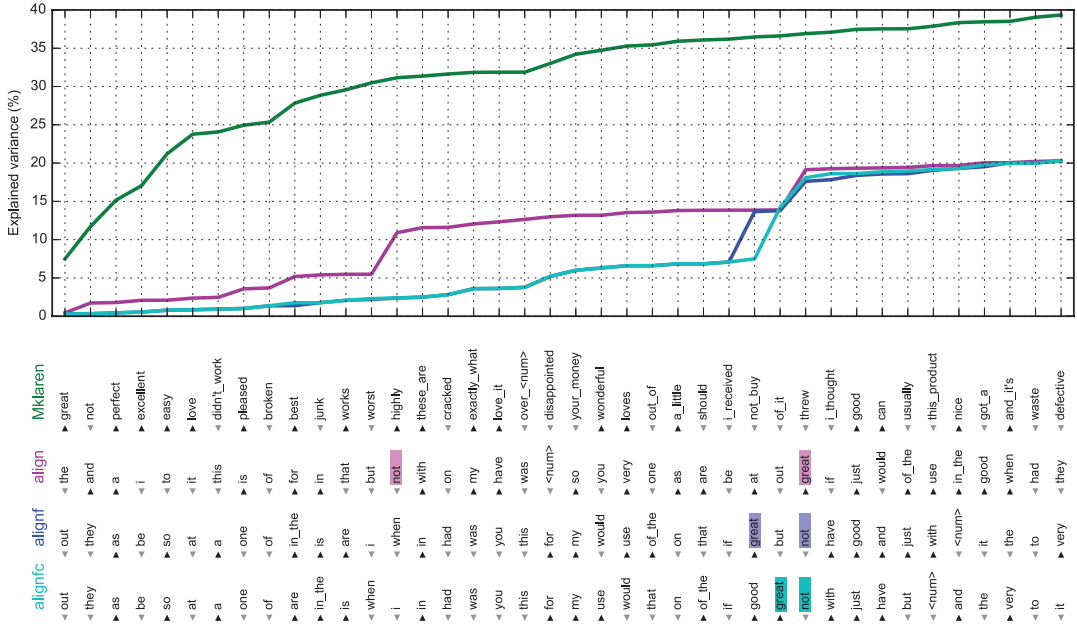}

\caption{
Increase in explained variance upon incrementally including features to an
ordinary least-squares model.  The order of features is determined by the
magnitude of kernel weights for \ctalign , \ctalignf\ and \ctalignfc\ or the
order of selection by \mklaren. Explained variance is measured as a ratio of
training RMSE vs. total variance. Arrows indicate positive
(black) or negative (gray) sign of the feature in the model weight vector upon
inclusion. Highlighted are words "great" and "not", which significantly alter
the explained variance when discovered by \ctalign , \ctalignf\ and \ctalignfc\ models.
}
\label{f:features_align}
\end{figure*}

Finally, we compare the methods with respect to feature selection on the
kitchen dataset. Each of the methods \mklaren, \ctalign, \ctalignf, and
\ctalignfc\ returns and ordering of the kernels (features).  With \mklaren, this
order is obtained as the pivot columns corresponding to kernels are added to
the approximation. With alignment-based methods, we use the order induced by
the kernel weight vector. In Fig.~\ref{f:features_align}, we display the top 40
features as obtained from each such ordering, shown as words on the horizontal
axis.  We incrementally add features to an \emph{active set}. As each feature
is added at step $i$, we infer an ordinary least-squares
$\mx{\beta}_{\text{OLS}, i}$, which uses all features up to $i$.  The explained
variance is calculated as the ratio of the difference of the RMSE on the
training set versus total variance. The arrows below each word at step $i$
indicate the sign of the corresponding weight in $\mx{\beta}_{\text{OLS}, i}$.

Intuitively, the slope (change in explained variance) is higher for features
corresponding to words associated to strong sentiments. This is most notable
for words such as \emph{great}, \emph{good}, \emph{love}, etc. Not
surprisingly, the order in which features are added to the model critically
influences the explained variance. Here, \mklaren\ outperforms the
alignment-based methods.  Due to its linear complexity in the number of kernels
$p$, the features strongly correlated to the response are identified early,
irrespective to their magnitude.  On the other hand, the centered alignment
appears to be biased towards words with a high number of nonzero entries in the
dataset, such as propositions. Moreover, the word associated to negative or
positive sentiments are approximately balanced, according to the signs in
$\mx{\beta}_{\text{OLS}, i}$. The results confirm \mklaren\ can be used for model
interpretation.

\section{Conclusion}

Subquadratic complexity in the number of training examples is essential in
large-scale application of kernel methods. Learning the kernel matrix
efficiently from the data and the selection of relevant portions on the data
early can reduce time and storage requirements further up the machine learning
pipeline. The complexity with respect to the number of kernels should not be
disregarded when the number of kernels is large. Using a greedy low-rank
approximation to multiple kernels, we achieve linear complexity in the number
of kernels and data points without sacrificing the consideration of in-between
kernel correlations.  Moreover, the approach learns a regression model, but is
nevertheless applicable in any kernel-based model. The extension to
classification or ranking tasks is an interesting subject for future work.
Contrary to the recent kernel matrix approximations, we present an idea based
entirely on geometric principles, which is not limited to transductive
learning.  With the abundance of different data representations, we expect
kernel methods to remain essential in machine learning applications.

\section*{Appendix}

\subsection*{Least-angle regression}
\label{app:lar}

Least-angle regression (LAR) is an \emph{active set method}, originally
designed for feature subset selection in linear
regression~\citep{friedman2001elements,Hesterberg2008, Efron2004}.  A column is
chosen from the set of candidates such that the correlations with the residual
are equal for all active variables.  This is possible because all variables
(columns) are known \emph{a priori}, which clearly does not hold for candidate
pivot columns. The monotonically decreasing maximal correlation in the active
set is therefore not guaranteed. Moreover, the addition of a column to the
active set potentially affects the values in all further columns.  Naively
recomputing these values at each iteration would yield a computational
complexity of order $O(n^2)$.

Let the predictor variables ${\mx{x}_1, \mx{x}_2, ..., \mx{x}_p}$ be vectors in
$\mathbb{R}^{n}$, arranged in a matrix $\mx{X} \in \mathbb{R}^{n \times p}$.
The associated response vector is $\mx{y} \in \mathbb{R}^{n}$.  The LAR method
iteratively selects the predictor variables $\mx{x}_j$ and the corresponding
coefficients $\beta_j$ are updated at the same time as they are moved towards
their least-squares coefficients. At last step, the method reaches the
least-squares solution $\mx{X}\mx{\beta} = \mx{y}$.

The high-level pseudo code is as follows:

\begin{enumerate}

\item Start with the residual $\mx{r} = \mx{y} - \bar{\mx{y}}$, and regression
coefficients $\beta_1, \beta_2, ... \beta_p = 0$.

\item Find the variable $\mx{x}_j$ most correlated with $\mx{r}$.

\item Move $\beta_j$ towards its least-squares coefficient until another
$\mx{x}_k$ has as much correlation with $\mx{r}$.

\item Move $\beta_j$ and $\beta_k$ in the direction towards their joint
least-sq. coeff., until some new $\mx{x}_l$ has as much correlation with
$\mx{r}$.

\item Repeat until all variables have been entered, reaching the least-sq.
solution.  

\end{enumerate}

Note that the method is easily modified to include early stopping, after a
maximum number of selected predictor variables are included.  Importantly, the
method can be viewed as a version of supervised Incomplete Cholesky
Decomposition of the \emph{linear kernel} $\mx{K} = \mx{X}\mx{X}^T$ which
corresponds to the usual inner product in $\mathbb{R}^p$.

Assume that the predictor variables are standardized and response has had its mean subtracted off:
\beq
\begin{split}
    \mx{1}^T \mx{x}_j &= 0 \text{ and }   \|\mx{x}_j\|_2 = 1 \text { for } j = 1, 2, ..., p . \\
    \mx{1}^T \mx{y} & = 0 
\end{split}
\eeq

Initialize the \emph{regression line} $\mx{\mu}$, the \emph{residual} \mx{r} and the
\emph{active set} $\mathcal{A}$: \beq
    \mx{\mu} = \mx{0} \text{,  } \mx{r} = \mx{y} \text{ and } \mathcal{A} = \emptyset \text{ .} 
\eeq

The LAR algorithm estimates $\mx{\mu} = \mx{X}\mx{\beta}$ in successive steps.
Say the predictor $\mx{x}_i$ has the largest correlation with $\mx{r}$. Then,
the index $i$ is added to the active set $\mathcal{A}$ and the regression line
and residual are updated:

\beq
\begin{split}
\mx{\mu}^{\text{new}} &= \mx{\mu} + \gamma \mx{x}_i \\
\mx{r}^{\text{new}} &= \mx{r} - \gamma \mx{x}_i
\end{split}
\label{se:lar_update}
\eeq

The step size $\gamma$ is set such that a new predictor $\mx{x}_j$ will enter
the model after $\mx{\mu}$ is updated and all predictors in the active set as
well as $\mx{x}_j$ will be equally correlated to \mx{r}. The key parts are the
selection of predictors added to the model and the calculation of the step size.

The active matrix for a subset of indices $j$ with sign $s_j$ is defined as
\beq
\begin{split}
\mx{X}_A &= \big( \cdots s_j \mx{x}_j \cdots \big)  \text{ for } j \in \mathcal{A} \\
s_j &= \text{sign}\{\mx{x}_j^T\mx{r}\}
\end{split}
\eeq

By elementary linear algebra, there exist a \emph{bisector} $\mx{u}_A$ - an
equiangular vector, having $\|\mx{u}_A\|_2 = 1$ and making equal angles, less
than 90 degrees, with vectors in $\mx{X}_A$. Define the following quantities
respectively: $\mx{X}_A$ the active matrix, $A$ the normalization scalar,
$\mx{u}_A$ the bisector, and \mx{\omega} the vector making equal angles with
the columns of $\mx{X}_A$. The bisector is obtained as follows.

\beq
\begin{split}
\mx{T}_A &= \mx{X}_A^T\mx{X}_A \\
A &= (\mx{1}_A^T\mx{T}_A\mx{1}_A)^{-1/2} \\
\mx{\omega} &= A \mx{T}_A^{-1}\mx{1}_A \\ 
\mx{u}_A &= \mx{X}_A \mx{\omega}_A 
\label{e:bisector}
\end{split}
\eeq

The calculation of step size $\gamma$ proceeds as follows. Get the maximum vector of
correlations. Active set contains variables with highest absolute correlations.

\beq
\begin{split}
c_j &= \mx{x}_j^T\mx{r} \\
C &= \text{max}_j\{c_j\} \\
\mx{a} &= \mx{X}_A^T \mx{u}_A \\
 \\
\gamma &= \text{min}^+_{j \in \mathcal{A}^c} \{\frac{C - c_j}{A_A - a_j}, \frac{C + c_j}{A_A + a_j}\}
\label{se:lar_minimum}
\end{split}
\eeq
 where $\text{min}^+$ is the minimum over positive components.  

By Eq.~\ref{se:lar_update}, we the change in correlations within the active set
can be expressed.
\beq
    c_j^{\text{new}} = \mx{x}_j^T(\mx{y} - \mx{r}^{\text{new}}) = c_j - \gamma a_j
    \label{se:lar_new_correlation}
\eeq
For the predictors in active set, we have
\beq
|c_j^{\text{new}}| = C - \gamma A ,  \text{  for } j \in \mathcal{A}.
\label{se:lar_new_correlation_a}
\eeq

A variable is selected from the remaining variables in $\mathcal{A}^c$, such
that $c_j^{\text{new}}$ is maximal.  Equaling Eq.~\ref{se:lar_new_correlation}
and Eq.~\ref{se:lar_new_correlation_a}, and maximizing yields $\gamma = \frac{C -
c_j}{A - a_j}$. Similarly, $-c_j^{\text{new}}$ for the reverse covariate is
maximal at $\gamma = \frac{C + c_j}{A + a_j}$. Hence, $\gamma$ is chosen in
Eq.~\ref{se:lar_minimum} as a minimal value for which an variable joins the
active set.

% \bibliographystyle{IEEEtran}
% \addcontentsline{toc}{section}{References}
\bibliography{ref}

\begin{thebibliography}{44}
\providecommand{\natexlab}[1]{#1}
\providecommand{\url}[1]{\texttt{#1}}
\expandafter\ifx\csname urlstyle\endcsname\relax
  \providecommand{\doi}[1]{doi: #1}\else
  \providecommand{\doi}{doi: \begingroup \urlstyle{rm}\Url}\fi

\bibitem[Bach(2012)]{Bach2012}
Francis Bach.
\newblock {Sharp analysis of low-rank kernel matrix approximations}.
\newblock \emph{arXiv:1208.2015}, August 2012.

\bibitem[Bach et~al.(2010)Bach, Mairal, Ponce, and Sapiro]{bach2010sparse}
Francis Bach, Julien Mairal, Jean Ponce, and Guillermo Sapiro.
\newblock {Sparse coding and dictionary learning for image analysis}.
\newblock In \emph{Proceedings of IEEE International Conference on Computer
  Vision and Pattern Recognition}, 2010.

\bibitem[Bach and Jordan(2005)]{Bach2005}
Francis~R. Bach and Michael~I. Jordan.
\newblock {Predictive low-rank decomposition for kernel methods}, 2005.

\bibitem[Bhattacharyya and Bhowmick(2015)]{Bhattacharyya2015}
Arnab Bhattacharyya and Abhishek Bhowmick.
\newblock {Pivoted Cholesky decomposition by Cross Approximation for efficient
  solution of kernel systems}.
\newblock \emph{arXiv preprint arXiv:1505.06195}, pages 1--19, May 2015.

\bibitem[Bishop(2006)]{bishop2006pattern}
Christopher~M Bishop.
\newblock \emph{{Pattern Recognition and Machine Learning}}.
\newblock Information Science and Statistics. Springer, 2006.
\newblock ISBN 9780387310732.

\bibitem[Blitzer et~al.(2007)Blitzer, Dredze, and
  Pereira]{blitzer2007biographies}
John Blitzer, Mark Dredze, and Fernando Pereira.
\newblock {Biographies, bollywood, boom-boxes and blenders: Domain adaptation
  for sentiment classification}.
\newblock In \emph{ACL}, volume~7, pages 440--447, 2007.

\bibitem[Cao et~al.(2015)Cao, Brubaker, Fleet, and Hertzmann]{Cao2015}
Yanshuai Cao, Marcus Brubaker, David Fleet, and Aaron Hertzmann.
\newblock {Efficient Optimization for Sparse Gaussian Process Regression}.
\newblock \emph{IEEE Transactions on Pattern Analysis \& Machine Intelligence},
  pages 1--1, 2015.
\newblock ISSN 0162-8828.
\newblock \doi{10.1109/TPAMI.2015.2424873}.

\bibitem[Cortes et~al.(2010)Cortes, Mohri, and Rostamizadeh]{cortes2010two}
Corinna Cortes, Mehryar Mohri, and Afshin Rostamizadeh.
\newblock {Two-stage learning kernel algorithms}.
\newblock In \emph{Proceedings of the 27th International Conference on Machine
  Learning}, pages 239--246, 2010.

\bibitem[Cortes et~al.(2012)Cortes, Mohri, and Rostamizadeh]{Cortes2012}
Corinna Cortes, Mehryar Mohri, and Afshin Rostamizadeh.
\newblock {Algorithms for Learning Kernels Based on Centered Alignment}.
\newblock \emph{Journal of Machine Learning Research}, 13:\penalty0 795--828,
  March 2012.

\bibitem[Efron and Hastie(2004)]{Efron2004}
Bradley Efron and Trevor Hastie.
\newblock {Least angle regression}.
\newblock \emph{The Annals of statistics}, 32\penalty0 (2):\penalty0 407--499,
  2004.

\bibitem[Fine and Scheinberg(2001)]{Fine2001}
Shai Fine and Katya Scheinberg.
\newblock {Efficient SVM Training Using Low-Rank Kernel Representations}.
\newblock \emph{Journal of Machine Learning Research}, 2:\penalty0 243--264,
  2001.

\bibitem[Friedman et~al.(2001)Friedman, Hastie, and
  Tibshirani]{friedman2001elements}
Jerome Friedman, Trevor Hastie, and Robert Tibshirani.
\newblock \emph{{The elements of statistical learning}}, volume~1.
\newblock Springer series in statistics Springer, Berlin, 2001.

\bibitem[Gal and Turner(2015)]{Gal2015}
Yarin Gal and R~Turner.
\newblock {Improving the Gaussian process sparse spectrum approximation by
  representing uncertainty in frequency inputs}.
\newblock In \emph{Proceedings of the 32nd International Conference on Machine
  Learning}, 2015.

\bibitem[Gittens and Mahoney(2013)]{Gittens2013}
Alex Gittens and Michael~W. Mahoney.
\newblock {Revisiting the Nystrom Method for Improved Large-Scale Machine
  Learning}.
\newblock \emph{arXiv preprint arXiv:1303.1849}, page~60, March 2013.

\bibitem[Golub and {Van Loan}(2012)]{golub2012matrix}
Gene~H Golub and Charles~F {Van Loan}.
\newblock \emph{{Matrix computations}}, volume~3.
\newblock JHU Press, 2012.

\bibitem[G\"{o}nen and Alpaydin(2010)]{Gonen2010}
Mehmet G\"{o}nen and Ethem Alpaydin.
\newblock {Supervised learning of local projection kernels}.
\newblock \emph{Neurocomputing}, 73\penalty0 (10-12):\penalty0 1694--1703, June
  2010.
\newblock ISSN 09252312.
\newblock \doi{10.1016/j.neucom.2009.11.043}.

\bibitem[G\"{o}nen and Alpaydin(2011)]{Gonen2011}
Mehmet G\"{o}nen and Ethem Alpaydin.
\newblock {Multiple kernel learning algorithms}.
\newblock \emph{The Journal of Machine Learning Research}, 12:\penalty0
  2211--2268, 2011.

\bibitem[Hesterberg et~al.(2008)Hesterberg, Choi, Meier, and
  Fraley]{Hesterberg2008}
Tim Hesterberg, Nam~Hee Choi, Lukas Meier, and Chris Fraley.
\newblock {Least angle and ℓ 1 penalized regression: A review}.
\newblock \emph{Statistics Surveys}, 2:\penalty0 61--93, 2008.
\newblock ISSN 1935-7516.
\newblock \doi{10.1214/08-SS035}.

\bibitem[Kandola et~al.(2002)Kandola, Shawe-Taylor, and
  Cristianini]{Kandola2002}
J~Kandola, J~Shawe-Taylor, and N~Cristianini.
\newblock {Optimizing Kernel Alignment over Combination of Kernels}, 2002.

\bibitem[Kaski and Gonen(2014)]{Kaski2014}
Samuel Kaski and Mehmet Gonen.
\newblock {Kernelized Bayesian matrix factorization}.
\newblock \emph{IEEE Transactions on Pattern Analysis \& Machine Intelligence},
  36\penalty0 (10):\penalty0 2047--2060, 2014.

\bibitem[Kulis et~al.(2009)Kulis, Sustik, and Dhillon]{Kulis2009}
Brian Kulis, MA~Sustik, and IS~Dhillon.
\newblock {Low-rank kernel learning with Bregman matrix divergences}.
\newblock \emph{The Journal of Machine Learning Research}, 10:\penalty0
  341--376, 2009.

\bibitem[Lanckriet and Cristianini(2004)]{Lanckriet2004a}
GRG Lanckriet and N~Cristianini.
\newblock {Learning the kernel matrix with semidefinite programming}.
\newblock \emph{Journal of Machine Learning Research}, 5:\penalty0 27--72,
  2004.

\bibitem[Le et~al.(2013)Le, Sarl\'{o}s, and Smola]{Le2013a}
Quoc Le, T~Sarl\'{o}s, and Alex Smola.
\newblock {Fastfood—approximating kernel expansions in loglinear time}.
\newblock \emph{Proceedings of the 30th International Conference on Machine
  Learning}, 28, 2013.

\bibitem[Li et~al.(2015)Li, Bi, Kwok, and Lu]{Li2015}
Mu~Li, Wei Bi, James~T Kwok, and Bao-Liang Lu.
\newblock {Large-Scale Nystr\"{o}m Kernel Matrix Approximation Using Randomized
  SVD}.
\newblock \emph{IEEE Transactions on Neural Networks and Learning Systems},
  26\penalty0 (1):\penalty0 152--164, 2015.

\bibitem[Meyer(2000)]{meyer2000matrix}
Carl~D Meyer.
\newblock \emph{{Matrix analysis and applied linear algebra}}.
\newblock Siam, 2000.

\bibitem[Mohsenzadeh et~al.(2015)Mohsenzadeh, Sheikhzadeh, and
  Member]{Mohsenzadeh2015}
Yalda Mohsenzadeh, Hamid Sheikhzadeh, and Senior Member.
\newblock {Gaussian Kernel Width Optimization for Sparse Bayesian Learning}.
\newblock \emph{IEEE Transactions on Neural Networks and Learning Systems},
  26\penalty0 (4):\penalty0 709--719, 2015.

\bibitem[Pennington and Yu(2015)]{Pennington}
Jeffrey Pennington and Felix~X Yu.
\newblock {Spherical Random Features for Polynomial Kernels}.
\newblock In \emph{Advances in Neural Information Processing Systems}, pages
  1837--1845, 2015.

\bibitem[Qui\~{n}onero Candela and Rasmussen(2005)]{Qui2005}
J~Qui\~{n}onero Candela and CE~Rasmussen.
\newblock {A unifying view of sparse approximate Gaussian process regression}.
\newblock \emph{The Journal of Machine Learning Research}, 6:\penalty0
  1939--1959, 2005.

\bibitem[Rahimi and Recht(2007)]{Rahimi2007}
Ali Rahimi and Ben Recht.
\newblock {Random features for large-scale kernel machines}.
\newblock In \emph{Advances in Neural Information Processing Systems}, pages
  1177--1184, 2007.

\bibitem[Rakotomamonjy and Chanda(2014)]{Rakotomamonjy2014}
Alain Rakotomamonjy and Sukalpa Chanda.
\newblock {ℓp-Norm Multiple Kernel Learning With Low-Rank Kernels}.
\newblock \emph{Neurocomputing}, 143:\penalty0 68--79, November 2014.
\newblock ISSN 09252312.
\newblock \doi{10.1016/j.neucom.2014.06.019}.

\bibitem[Rudi et~al.(2015)Rudi, Camoriano, and Rosasco]{rudi2015less}
Alessandro Rudi, Raffaello Camoriano, and Lorenzo Rosasco.
\newblock {Less is More: Nystrom Computational Regularization}.
\newblock \emph{arXiv preprint arXiv:1507.04717}, 2015.

\bibitem[Sch\"{o}lkopf and Smola(2002)]{scholkopf2002learning}
Bernhard Sch\"{o}lkopf and Alexander~J Smola.
\newblock \emph{{Learning with kernels: Support vector machines,
  regularization, optimization, and beyond}}.
\newblock MIT press, 2002.

\bibitem[Si et~al.(2014)Si, Hsieh, and Dhillon]{Si2014}
Si~Si, Cho-Jui Hsieh, and Inderjit Dhillon.
\newblock {Memory Efficient Kernel Approximation}.
\newblock \emph{Proceedings of The 31st International Conference on Machine
  Learning}, 32, 2014.

\bibitem[Snelson and Ghahramani(2006)]{Snelson2006}
Edward Snelson and Zoubin Ghahramani.
\newblock {Sparse Gaussian processes using pseudo-inputs}.
\newblock In \emph{Proceedings of the 23nd international conference on Machine
  learning - ICML '06}, 2006.

\bibitem[Sonnenburg et~al.(2005)Sonnenburg, R\"{a}tsch, and
  Sch\"{o}lkopf]{sonnenburg2005large}
S\"{o}ren Sonnenburg, Gunnar R\"{a}tsch, and Bernhard Sch\"{o}lkopf.
\newblock {Large scale genomic sequence SVM classifiers}.
\newblock In \emph{Proceedings of the 22nd international conference on Machine
  learning}, pages 848--855. ACM, 2005.

\bibitem[Szabo(2015)]{Szabo2015}
Zoltan Szabo.
\newblock {Optimal rates for Random Fourier Features}.
\newblock In \emph{Advances in Neural Information Processing Systems}, pages
  1144--1152, May 2015.

\bibitem[Vedaldi and Zisserman(2012)]{Vedaldi2012}
Andrea Vedaldi and Andrew Zisserman.
\newblock {Efficient additive kernels via explicit feature maps.}
\newblock \emph{IEEE Transactions on Pattern Analysis \& Machine Intelligence},
  34\penalty0 (3):\penalty0 480--92, March 2012.
\newblock ISSN 1939-3539.
\newblock \doi{10.1109/TPAMI.2011.153}.

\bibitem[Williams and Seeger(2001)]{Williams2001}
Christopher Williams and Matthias Seeger.
\newblock {Using the Nystr\{\"{o}\}m method to speed up kernel machines}.
\newblock In \emph{Proceedings of the 14th Annual Conference on Neural
  Information Processing Systems}, pages 682--688, 2001.

\bibitem[Wilson(2015)]{Wilson2015}
Andrew~Gordon Wilson.
\newblock {Kernel Interpolation for Scalable Structured Gaussian Processes (
  KISS-GP )}.
\newblock \emph{arXiv preprint arXiv:1503.01057}, 37, 2015.

\bibitem[Wilson and Adams(2013)]{Wilson2013}
Andrew~Gordon Wilson and Ryan Adams.
\newblock {Gaussian process kernels for pattern discovery and extrapolation}.
\newblock \emph{arXiv preprint arXiv:1302.4245}, 28, 2013.

\bibitem[Xu et~al.(2015)Xu, Jin, Shen, and Zhu]{Xu2015}
Zenglin Xu, Rong Jin, Bin Shen, and Shenghuo Zhu.
\newblock {Nystrom Approximation for Sparse Kernel Methods: Theoretical
  Analysis and Empirical Evaluation}.
\newblock In \emph{Twenty-Ninth AAAI Conference on Artificial Intelligence},
  pages 3115--3121, 2015.

\bibitem[Yang et~al.(2014)Yang, Smola, Song, and Wilson]{Yang2014}
Zichao Yang, Alexander~J. Smola, Le~Song, and Andrew~Gordon Wilson.
\newblock {A la Carte - Learning Fast Kernels}.
\newblock \emph{arXiv preprint arXiv:1412.6493}, December 2014.

\bibitem[Zhang et~al.(2012)Zhang, Wang, and Moerchen]{Zhang2012}
Kai Zhang, Zhuang Wang, and Fabian Moerchen.
\newblock {Scaling up Kernel SVM on Limited Resources : A Low-rank
  Linearization Approach}.
\newblock In \emph{International Conference on Artificial Intelligence and
  Statistics}, volume~XX, pages 1425--1434, 2012.

\bibitem[Zou and Hastie(2005)]{Zou2005}
Hui Zou and Trevor Hastie.
\newblock {Regularization and variable selection via the elastic net}.
\newblock \emph{Journal of the Royal Statistical Society: Series B (Statistical
  Methodology)}, 67\penalty0 (2):\penalty0 301--320, April 2005.
\newblock ISSN 1369-7412.
\newblock \doi{10.1111/j.1467-9868.2005.00503.x}.

\end{thebibliography}

\end{document}